\documentclass[numsec,webpdf,modern,large,round]{oup-authoring-template}
\usepackage{subfig}
\usepackage{float}	%用于排版图片位置
\usepackage{hyperref}
\usepackage{booktabs}
\usepackage[utf8]{inputenc}
\usepackage{graphicx} % Required for inserting images
\usepackage{lmodern} % Load Latin Modern fonts
\usepackage{fix-cm} % Fix Computer Modern fonts
\usepackage{multirow} % Required for multirow cells, if needed
\usepackage{siunitx}  % For aligning numbers by decimal point and handling +/-
\usepackage{caption}  % For table captions
\theoremstyle{thmstyleone}%
\theoremstyle{thmstyletwo}%
\theoremstyle{thmstylethree}%

\begin{document}

\journaltitle{Bioinformatics}
% \DOI{DOI HERE}
\copyrightyear{2025}
% \pubyear{2019}
% \access{Advance Access Publication Date: Day Month Year}
% \appnotes{Paper}

% \journaltitle{}  % 设置为空字符串
% \DOI{}  % 设置为空字符串
% \copyrightyear{}  % 设置为空字符串

\firstpage{1}

%\subtitle{Subject Section}

\title[]{BioGraphFusion: Graph Knowledge Embedding for Biological Completion and Reasoning}
%%BioGraphFusion: Graph Knowledge Embedding for Completion and Reasoning over Biological Knowledge Graphs
% \title[Short Article Title]{BioGraphFusion: A Graph Neural Network and Knowledge Embedding Framework for Completion and Reasoning over Biological Knowledge Graphs}
% BioGraphCR: A Graph Neural Network and Knowledge Embedding Framework for Completion and Reasoning over Biological Knowledge Graphs
% KGNet-Fusion: Integrating Graph Structure and Knowledge Embedding for Enhanced Biological Knowledge Graph Reasoning
% SemGraph-KG: Semantic-Driven Graph Neural Network for Biological Knowledge Graph Completion and Reasoning
\author[1,$\dagger$]{Yitong Lin}
\author[1,$\dagger$]{Jiaying He}
\author[1]{Jiahe Chen}
\author[1]{Xinnan Zhu}
\author[1,$\ast$]{Jianwei Zheng}
\author[2,$\ast$]{Tao Bo}
% \author[3]{Fourth Author}
% \author[4]{Fifth Author\ORCID{0000-0000-0000-0000}}

\authormark{Lin et al.}

\address[1]{\orgdiv{College of Computer Science and Technology}, \orgname{Zhejiang University of Technology}, \orgaddress{\street{Hangzhou},  \state{Zhejiang}, \postcode{310023}, \country{China}}}

\address[2]{\orgdiv{Key Laboratory of Endocrine Glucose \& Lipids Metabolism},% and Brain Aging, Ministry of Education
\orgname{Department of Endocrinology, Shandong Provincial Hospital Affiliated to Shandong First Medical University},
\orgaddress{ \street{Jinan}, \state{Shandong}, \country{China}}    }

%\address[3]{\orgdiv{Central Laboratory},
%\orgname{Shandong Provincial Hospital Affiliated to Shandong First Medical University},
%\orgaddress{\street{Jinan}, \state{Shandong}, \postcode{250021},  \country{China}}}
% \address[4]{\orgdiv{Department}, \orgname{Organization}, \orgaddress{\street{Street}, \postcode{Postcode}, \state{State}, \country{Country}}}

\corresp[$\dagger$]{Equal contributions.}

\corresp[$\ast$]{Corresponding author. }

\received{Date}{0}{Year}
\revised{Date}{0}{Year}
\accepted{Date}{0}{Year}

%\editor{Associate Editor: Name}

\abstract{
\textbf{Motivation:} 
Biomedical knowledge graphs (KGs) are crucial for drug discovery and disease understanding, yet their completion and reasoning are challenging. Knowledge Embedding (KE) methods capture global semantics but struggle with dynamic structural integration, while Graph Neural Networks (GNNs) excel locally but often lack semantic understanding. Even ensemble approaches, including those leveraging language models, often fail to achieve a deep, adaptive, and synergistic co-evolution between semantic comprehension and structural learning. Addressing this critical gap in fostering continuous, reciprocal refinement between these two aspects in complex biomedical KGs is paramount.\\
\textbf{Results:} We introduce BioGraphFusion, a novel framework for deeply synergistic semantic and structural learning. BioGraphFusion establishes a global semantic foundation via tensor decomposition, guiding an LSTM-driven mechanism to dynamically refine relation embeddings during graph propagation. This fosters adaptive interplay between semantic understanding and structural learning, further enhanced by query-guided subgraph construction and a hybrid scoring mechanism. Experiments across three key biomedical tasks demonstrate BioGraphFusion's superior performance over state-of-the-art KE, GNN, and ensemble models. A case study on Cutaneous Malignant Melanoma 1 (CMM1) highlights its ability to unveil biologically meaningful pathways.\\
\textbf{Availability and Implementation:} Source code and all training data are freely available for download at \url{https://github.com/Y-TARL/BioGraphFusion}.\\
\textbf{Contact:} \href{mailto:zjw@zjut.edu.cn}{zjw@zjut.edu.cn}, \href{mailto:botao666666@126.com}{botao666666@126.com}
\\
\textbf{Supplementary information:} Supplementary data are available at \textit{Bioinformatics}
online.}


% \boxedtext{
% \begin{itemize}
% \item Key boxed text here.
% \item Key boxed text here.
% \item Key boxed text here.
% \end{itemize}}

\maketitle

\section{Introduction}
Knowledge Graphs (KGs) are semantic networks that represent relationships between entities as a set of triples \((h, r, t)\), where \(h\) and \(t\) denote the head and tail entities, and \(r\) represents the relation connecting them ~\citep{kdgene}. These graphs model real-world concepts and their interactions through nodes (entities) and edges (relations). Specifically, biological knowledge graphs have extended this framework to encompass entities such as diseases, genes, drugs, chemicals, and proteins, facilitating a structured understanding of clinical knowledge.

Technically, large-scale biological KGs such as DisGeNET~\citep{DisGeNET}, STITCH~\citep{stitch}, and SIDER~\citep{sider} are widely used in biomedical research, which support applications including disease gene prediction~\citep{KGEPredicting_Disease-Gene_Associations}, drug-target interaction~\citep{causal}, and drug-drug correlation~\citep{knowddi}. 
Many such tasks demand for practical techniques of Knowledge Graph Completion (KGC) ~\citep{KnowledgeGraphCompletion} and Knowledge Graph Reasoning (KGR) ~\citep{KnowledgeGraphReasoning}. Fundamentally, both techniques involve predicting the answer to a query of the form (h, r, ?)~\citep{Structure-Information-Based_Reasoning}, to identify the missing tail entity. While both may be considered as link prediction, they differ: KGC primarily predicts missing direct links (entities or relations) by identifying patterns in existing graph data. Extending this, KGR is a broader task that infers complex or multi-step knowledge, often employing logical inference mechanisms, rule-based systems, or multi-hop path analysis to deduce unstated facts. Thus, KGC focuses on completing the KG based on observed patterns, while KGR derives new insights through deeper inferential processes.

For KGC and KGR, Knowledge Embedding (KE) and Graph Structure Propagation (GSP) are prevalent approaches, as detailed in foundational works~\citep{tang2024fusing,KnowledgeGraphReasoning}. KE, often termed a latent feature model~\citep{reviewofkg}, embeds entities and relations into continuous vector spaces, capturing semantic information to score candidate entities directly. For instance, RotatE~\citep{rotate} models relations as rotations in complex space \((h \circ r \approx t\)), while CP-N3~\citep{cp-n3} enhances performance by factorizing higher-order interactions. While KE techniques excel at capturing semantics, they often overlook structural patterns—such as multi-hop paths—limiting their reasoning capabilities over complex, multi-relational biomedical graphs~\citep{peng2023knowledge, liu2022learning}.

In contrast, GSP borrows its main architecture from Graph Neural Networks (GNNs) ~\citep{KEGCN}, which have significantly advanced network analysis by propagating messages between entities, thereby partially capturing topological information. Representatives such as GNN4DM ~\citep{GNN4DM} demonstrate the power of these approaches in tasks like discovering overlapping functional disease modules through the integration of network topology and genomic data. However, these methods, while adept at structural modeling, often tend to overemphasize topological information at the expense of deeper semantic associations and the rich content of relations.
Recognizing the limitations of traditional KE and GSP methods, and with the advent of powerful pre-trained language models (LMs), more advanced approaches have emerged for KGC and KGR. These methods often seek to incorporate richer semantic understanding directly from textual data or find novel ways of integrating semantic and structural information, moving beyond the KE or GSP paradigms alone.
% For instance, KG-BERT~\citep{KG-BERT} primarily employs pre-trained language models to score textualized triples, focusing on deep semantic comprehension yet suffering considerable computational cost. Other approaches, such as LASS~\citep{LASS}, aim for an explicit fusion by jointly embedding natural language semantics from triplet descriptions with graph structural information through fine-tuning LMs, using a probabilistic loss that reconstructs structural patterns. However, this interaction, largely mediated by the loss function, may not fully facilitate a deeply adaptive interplay.
For instance, KG-BERT~\citep{KG-BERT}  uses pre-trained LMs to score textualized triples, prioritizing semantics but with high computational costs. Similarly, LASS~\citep{LASS} attempts fusion by embedding natural language semantics with graph structure through LM fine-tuning and probabilistic reconstruction loss, yet its loss-mediated interaction limits deeply adaptive integration.

% These observations highlight the complementary strengths and limitations of KE and GSP, motivating the design of BioGraphFusion—a novel framework that synergistically integrates these two counterparts, striving for the joint optimization of node and relation embeddings. 

% While prior efforts advanced the integration of semantic and structural learning for KGC and KGR, they also reveal a persistent challenge: achieving a more profound coupling between semantic guidance and structural propagation for synergistic models. This critical gap—the need for a framework that fosters reciprocal and adaptive refinement between semantic understanding and structural learning—motivates our design of BioGraphFusion. BioGraphFusion is a novel framework conceived to synergistically integrate these semantic and structural paradigms. It strives for the joint and iterative optimization of both node and relation embeddings, thereby designed to mitigate the limitations observed in previous hybrid methodologies.
While these approaches significantly advanced semantic and structural learning for KGC and KGR, they underscore a persistent challenge: achieving deep, dynamic coupling where semantic guidance and structural propagation synergistically co-evolve. Many methods, despite innovations, still struggle with fully reciprocal, adaptive refinement between rich semantic understanding and nuanced structural learning. This critical gap—the difficulty in developing a framework that enables a mutually enhancing co-evolution between semantic and structural learning—motivates BioGraphFusion. BioGraphFusion is a novel framework designed for such a profound synergistic integration, which leverages semantic insight, primarily drawing from principles of KE for global context, and combines it with dynamic structural reasoning, inspired by GSP techniques. The overall goal is for joint optimization of node and relation embeddings, thereby addressing the limitations in achieving the deep and adaptive semantic-structural interplay seen in prior methods.

BioGraphFusion actualizes this for biomedical KGs by weaving global semantic modeling with dynamic structural reasoning. Initially, a Canonical Polyadic (CP) decompositio~\citep{kolda2009tensor} module establishes a global semantic foundation, extracting low-dimensional embeddings capturing overarching biological associations and cross-domain interactions. This global semantic framework then actively steers structural learning. An LSTM-based gating mechanism dynamically refines relation embeddings during message propagation, adapting them to evolving semantic contexts and enabling the model to better capture long-range dependencies crucial for complex biological pathways. Further, a query-guided subgraph construction component focuses structural exploration on pertinent biological regions, ensuring message passing and representation learning concentrate on relevant interactions. Finally, a hybrid scoring mechanism orchestrates synergy between these semantic and structural representations. Such balanced integration empowers the semantic model to guide dynamic refinement of graph-based representations, fostering deep optimization of intricate edge embeddings. 
This process ensures a reciprocal and adaptive refinement cycle, where semantic understanding and structural learning iteratively enhance each other.

% The main difference between KE and NE is that
% the latter focuses on the topology of the network, while KE
% focuses more on the internal information of different relations
% and the semantic connotation of facts.
% \vspace{-2em}
\section{Materials and methods}
\subsection{Dataset Overview and Task Design}
To advance biological knowledge graph completion and reasoning, we introduce three tasks integrating multi-source biomedical data.
First, the Disease-Gene Association Prediction task~\citep{kdgene} identifies missing disease-related genes by leveraging a primary dataset enriched with drug-disease and protein-chemical information.
Second, the Protein-Chemical Interaction task focuses on identifying compounds that interact with specific proteins, using core interaction data supplemented by auxiliary associations. 
Finally, the Cross-Medical Ontology Reasoning task employs the UMLS Terminology~\citep{UMLS}. This task functions as link prediction: given a head concept and relation type, the model predicts the tail concept, inferring diverse ontological relationships, including hierarchical and associative links.
Detailed dataset statistics and integration protocols for all tasks are summarized in Supplemental Materials Section 1 (SM1). 

% To evaluate the performance of our model on capabilities of biological completion and reasoning, we design three practical applications that integrate multi-source background knowledge.\\
% \textbf{Disease-Gene Association Prediction} 
%    The task involves predicting missing disease-related genes by leveraging a primary dataset of disease-gene associations (from DisGeNET~\citep{DisGeNET}) and complementary drug-disease and protein-chemical databases. This integration enables the inference of both direct and indirect associations.\\
% \textbf{Protein-Chemical Interaction Prediction}
% This task aims to identify compounds that interact with proteins by predicting their functional bindings. It is supported by a primary dataset of protein-chemical interactions (from STITCH~\citep{stitch}), along with auxiliary information derived from drug-disease and disease-gene interactions.\\
% \textbf{Cross-Medical Ontology Reasoning}
%    To assess multi-hop reasoning in complex medical ontologies, we employ the UMLS Terminology~\citep{UMLS}, which provides hierarchical relationships among medical terms such as organs, functions, and derived diseases.
% A complete summary of the datasets, including source details and statistics, is presented in Table~\ref{dataset} in the Experiments section.

% 注释图表
\begin{figure*}[htbp]%调节图片位置，h：浮动；t：顶部；b:底部；p：当前位置
\centering
% {\color{black!20}\rule{213pt}{37pt}}
 \includegraphics[width=.99\textwidth]{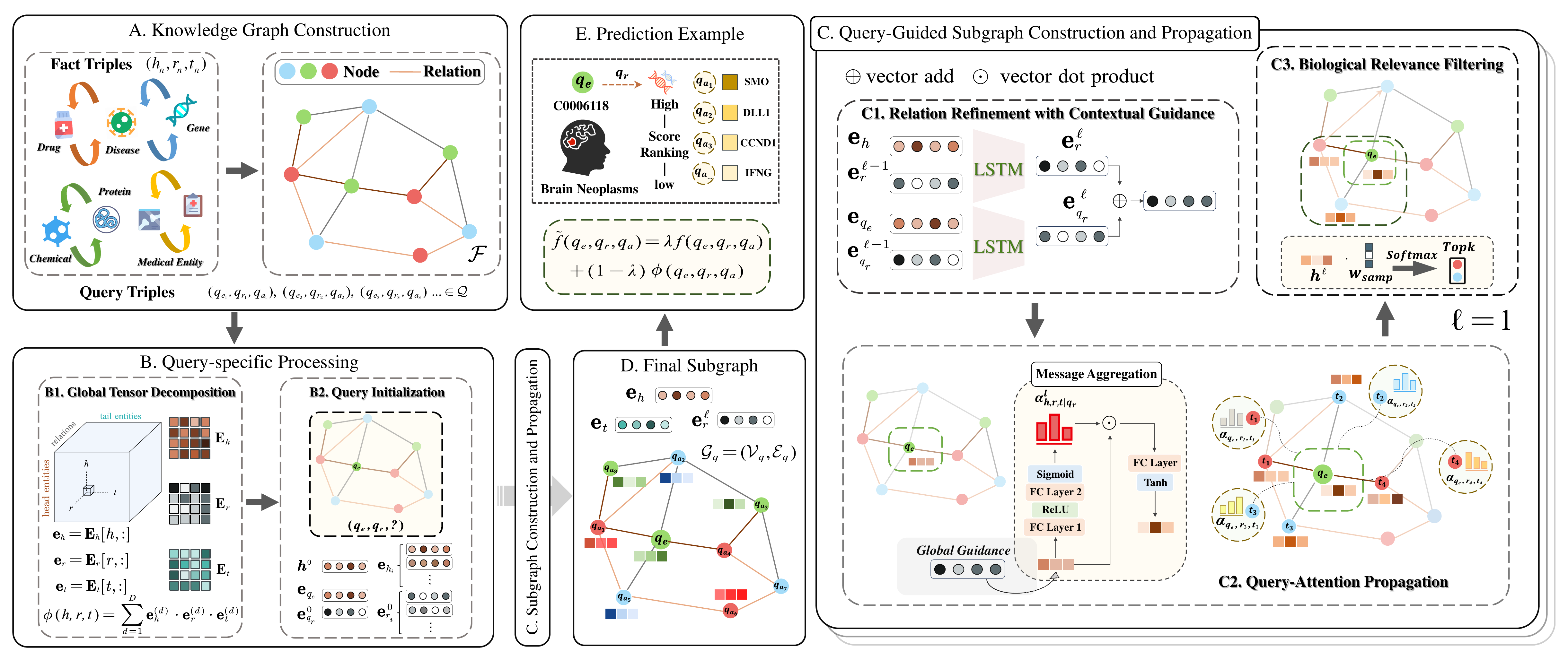}
\caption{Overview of the BioGraphFusion framework. (A) Knowledge Graph Construction: Integrating biomedical datasets to form a unified knowledge graph for downstream tasks. (B) Query-Specific Processing: A two-step process involving (B1) Global Tensor Decomposition that captures latent biological associations, and (B2) Query Initialization that guides the guide the subsequent process. (C) Subgraph Construction and Propagation: Iteratively builds a query-relevant
subgraph through neighborhood expansion and propagation, including (C1) Relation Refinement via LSTM, (C2) Query-Attention Propagation with context-based attention weights, and (C3) Biological Relevance Filtering to select the most pertinent entities. (D) Final Subgraph. (E) Scoring Integration that balances structural-semantic information and Prediction Example that selects the most promising predictions, with a focus on Brain Neoplasms.}\label{fig:overall}
\vspace{-1em}
\end{figure*}
\vspace{-1em}
\subsection{Overview of BioGraphFusion}  
BioGraphFusion achieves high performance in biomedical completion and reasoning by fostering a deep synergistic interplay between Knowledge Embedding (KE) and Graph Structure Propagation (GSP) principles.
By incorporating global semantic knowledge from knowledge embeddings to guide the graph propagation process, our proposal effectively captures both direct and long-range relationships in biomedical graphs.

As illustrated in Fig.~\ref{fig:overall}, BioGraphFusion comprises three key components. First, Global Biological Tensor Encoding (Section~\ref{sec:tensor}) employs Canonical Polyadic decomposition to extract low-dimensional embeddings that encode latent biological associations. Second, Query-Guided Subgraph Construction and Propagation (Section~\ref{sec:subgrah}) iteratively builds a query-relevant subgraph by refining relations and propagating context-specific embeddings.
Finally, these complementary aspects are unified through a hybrid scoring mechanism (Section~\ref{sec:score}). This mechanism integrates KE's direct global semantic contributions with structural insights from the KE-informed GSP process, enabling a nuanced assessment of candidate predictions.

\subsubsection{Notations and Problem Setup}
Let \(G = (\mathcal{V}, \mathcal{R}, \mathcal{F}, \mathcal{Q})\) be a biomedical knowledge graph integrating diverse fact triples from multiple sources for various tasks, as shown in Fig.~\ref{fig:overall}(A). Here, \( \mathcal{V} \) is the set of entities and \( \mathcal{R} \) the set of relations. \( \mathcal{F} \) is the set of factual triples, \( \mathcal{F} = \{ (h, r, t) \mid h, t \in \mathcal{V}, r \in \mathcal{R} \} \), where head entity \( h \) and tail entity \( t \) are connected by relation \( r \). To enhance graph diversity and model robustness, we also incorporate triples with reverse and identity relations ~\citep{redgnn,adaprop}. \( \mathcal{Q} \) contains query triples, \( \mathcal{Q} = \{ (q_e, q_r, q_a) \mid q_e, q_a \in \mathcal{V}, q_r \in \mathcal{R} \} \). Each query is of the form \( (q_e, q_r, ?) \), with \( q_a \) as the unknown target entity. The objective for such queries is to predict \( q_a \), a task central to KGC and KGR aimed at enriching the KG.

\vspace{-1em}
\subsubsection{Tensor Decomposition and Query Initialization}\label{sec:tensor}

% In biomedical knowledge analysis, capturing the global semantic landscape is crucial for effective reasoning and accurate predictions. Traditional methods that rely solely on local graph structures or randomly initialized embeddings often fail to encode the complex, latent relationships within a knowledge graph, leading to suboptimal performance—particularly for tasks requiring deep reasoning over interconnected biological entities. 

Effective biomedical knowledge analysis hinges on understanding the global semantic landscape to foster a dynamic interplay between semantic insights and structural patterns.
BioGraphFusion initiates this by establishing a global semantic foundation through tensor decomposition of the entire knowledge graph. This initial step provides a rich context essential for the subsequent integration of structural patterns with semantic understanding.
For this critical stage, we employ Canonical Polyadic (CP) decomposition~\citep{kolda2009tensor}. CP is chosen as it directly factorizes the graph's adjacency tensor to derive meaningful, low-dimensional latent embeddings for entities and relations. This factorization process adeptly captures fundamental relationships. Moreover, CP's formulation as a low-rank tensor approximation offers a balance between model expressiveness and parsimony, ensuring computational efficiency and scalability vital for processing large-scale biomedical knowledge graphs.

The graph tensor \(\mathcal{T} \in \mathbb{R}^{|\mathcal{V}| \times |\mathcal{R}| \times |\mathcal{V}|}\) is factorized via CP into three matrices: \(\mathbf{E}_h \in \mathbb{R}^{|\mathcal{V}| \times D}\), \(\mathbf{E}_r \in \mathbb{R}^{|\mathcal{R}| \times D}\), and \(\mathbf{E}_t \in \mathbb{R}^{|\mathcal{V}| \times D}\). These matrices capture the latent semantic associations between entities and relations (Fig.~\ref{fig:overall}~(B1)). The compatibility of any triple \((h, r, t)\) is then computed as:
\begin{equation} \phi(h, r, t) = \sum_{d=1}^D \mathbf{e}_h^{(d)} \cdot \mathbf{e}_r^{(d)} \cdot \mathbf{e}_t^{(d)} \label{phi} \end{equation}
where \(\mathbf{e}_h^{(d)}\), \(\mathbf{e}_r^{(d)}\), and \(\mathbf{e}_t^{(d)}\) are the \(d\)-th components of the respective embeddings.
 
Subsequently, BioGraphFusion initializes query-specific repres-entations directly from the CP-extracted matrices, ensuring a semantically meaningful starting point. Specifically, given a query \((q_e, q_r, ?)\), the entity embedding \(\mathbf{e}_{q_e}\) and the initial relation embedding \(\mathbf{e}_{q_r}^{0}\) are retrieved from \(\mathbf{E}_h\) and \(\mathbf{E}_r\), respectively (Fig.~\ref{fig:overall}\,(B2)). The initial node representation \(h^0\) is thus set to \(\mathbf{e}_{q_e}\), establishing a query-grounded context before neighborhood expansion. Similarly, all entity and relation embeddings in the graph, including those encountered during propagation, are initialized from CP decomposition, preserving global structural information for subgraph construction and message passing. 
\vspace{-1em}

\subsubsection{Query-Guided Subgraph Construction} \label{sec:subgrah}

Biomedical knowledge graphs are vast and noisy, making it computationally impractical and error-prone to process the entire graph for each query. To address this, we employ a query-guided subgraph construction mechanism that selectively expands along semantically relevant paths (see Fig.~\ref{fig:overall}~(C)), ensuring biological meaningfulness while filtering out spurious connections. 
% This adaptive approach focuses computational resources on the most relevant portions of the graph, enhancing both efficiency and interpretability in biomedical completion and reasoning tasks.

% % 注释图表
% \begin{figure}[htbp]%调节图片位置，h：浮动；t：顶部；b:底部；p：当前位置
% \centering
% % {\color{black!20}\rule{213pt}{37pt}}
%  \includegraphics[width=.5\textwidth]{Figures/propagation.pdf}
% \caption{Overview of the Subgraph Construction and Propagation, which iteratively builds a query-relevant subgraph through neighborhood expansion and propagation, including (C1) Relation Refinement via LSTM, (C2) Query-Attention Propagation with context-based attention weights, and (C3) Biological Relevance Filtering to select the most pertinent entities.}\label{fig:propagation}
% \end{figure}

\textbf{Neighborhood Expansion}
At each layer \( \ell \), the model expands the neighborhood for further propagation. Initially, at \( \ell = 0 \), the entity set \(\mathcal{V}^{(0)}\) contains only the query node \( q_e \). For each entity \( h \) at layer \( \ell - 1 \), we construct the candidate set \( \mathcal{C}^{(\ell)} \) by aggregating all direct neighbors of the current nodes:
\begin{equation}
\mathcal{C}^{(\ell)} = \bigcup_{h \in \mathcal{V}^{(\ell-1)}} \{ t \mid (h, r, t) \in \mathcal{F} \}
\end{equation}

%This candidate set \( \mathcal{C}^{(\ell)} \) includes all potential neighbors of the current entities. 
In this step, the model gathers all possible entities that can serve as neighbors for the current nodes during propagation.
On that basis, standard GNNs often update each node representation iteratively by gathering information from the surrounding entities. We also follow this step in our approach to constructing a candidate set $\mathcal{C}^{(\ell)}$ to prepare for message propagation. 
% However, before updating the node representations, we consider that it is crucial to refine the relation embeddings that capture the evolving semantic context, whose description is given in the next subsection.

\textbf{Contextual Relation Refinement} In many existing approaches, relation embeddings remain static or minimally updated, failing to account for contextual variations. However, in biomedical knowledge graphs, relations are rarely fixed; their meaning is shaped by the entities involved and the reasoning path. For instance, the relation “disease\_gene” can imply different biological mechanisms depending on the specific genes or proteins connected. Furthermore, static embeddings struggle to model multi-step interactions, such as indirect associations mediated by proteins or chemicals.

To mitigate these limitations, we introduce a Contextual Relation Refinement (CRR) module. LSTMs are chosen for their stateful transformation and gating mechanisms, which allow them to effectively model how relation meanings vary with entity context—a common scenario in biomedical KGs. Unlike simpler recurrent units (e.g., RNNs, GRUs), LSTMs excel at refining relation representations based on evolving semantic contexts from connected entities. This yields context-specific embeddings better suited for the dynamic, multi-step nature of biomedical relationships, iteratively updating relation embeddings as well as capturing context-dependent semantics and long-range dependencies~\citep{kdgene}. Specifically, for each triple \((h, r, t)\), the LSTM updates the relation embedding \(\mathbf{e}_r^{\ell}\) at layer \(\ell\), using the previous embedding \(\mathbf{e}_r^{\ell-1}\) as input and the head entity embedding \(\mathbf{e}_h\) as the hidden state:
\begin{equation}
\mathbf{e}_r^{\ell} = \text{LSTM}~(\mathbf{e}_r^{\ell-1}, \mathbf{e}_h)
\label{lstmr}
\end{equation}

Through the internal gating mechanisms (including the forget gate \(f\), input gate \(i\), candidate memory cell \(\tilde{c}\), memory cell \(c\), and output gate \(o\)), LSTM selectively processes and retains relevant contextual information. It tailors the relation embedding to the connected entities. Similarly, the query relation \(\mathbf{e}_{q_r}^{\ell}\) is updated by:
\begin{equation}
\mathbf{e}_{q_r}^{\ell} = \text{LSTM}~(\mathbf{e}_{q_r}^{\ell-1}, \mathbf{e}_{q_e})
\label{lstmqr}
\end{equation}
% The dual LSTM refines both head-node and query relations by adaptively adjusting their representations based on the semantic context of their respective entity embeddings. This dynamic modulation helps the model understand nuanced relationships. \textcolor{red}{Comparative experiments (see SM8 for details) confirmed LSTMs surpassed simpler alternatives, validating their selection for achieving the deep semantic-structural coupling central to our model.}
The dual LSTM adaptively refines both head-node and query relation representations based on semantic context from their respective entity embeddings. This dynamic modulation helps the model grasp nuanced relationships. Comparative experiments (see SM8 for details) have confirmed that LSTMs are better than other alternatives, validating the capability of achieving deep semantic-structural coupling central to our model.

\textbf{Query-Attention Propagation} 
Inspired by RED-GNN~\citep{redgnn}, each candidate node \( t \) aggregates messages from its neighbors using a query-attentive mechanism (Fig.~\ref{fig:overall}~(C2)). 
%Initially, since the only candidate node is \( q_e \), its representation \( \mathbf{h}^0(q_e, q_r) \) is directly initialized from the CP-extracted matrices(i.e., $\mathbf{e}_{q_e}$). 
% At layer \( \ell \), node representations are updated as  
Specifically, the node representation at layer \( \ell \) is updated as
\begin{equation*}
\mathbf{h}_t^\ell(q_e, q_r) = \delta \!\left( \mathbf{W}^\ell \cdot \sum_{(h, r, t) \in \mathcal{C}^{(\ell)}} \alpha_{h,r,t|q_r}^\ell \left( \mathbf{h}_h^{\ell-1}(q_e, q_r) + \mathbf{e}_r^\ell \right) \right)
\end{equation*}
where \( \mathbf{W}^\ell \) is a trainable weight matrix and \( \delta \) denotes the Tanh activation function. The attention weight \( \alpha_{h,r,t|q_r}^\ell \), computed as

\begin{equation*}
\alpha_{h,r,t|q_r}^\ell = \sigma \!\left( \left( \mathbf{w}_\alpha^\ell \right)^\top \text{ReLU}\!\left( \mathbf{W}_\alpha^\ell \cdot \left[ \mathbf{h}_h^{\ell-1}(q_e, q_r) + \mathbf{e}_r^\ell + \mathbf{e}_{q_r}^\ell \right] \right) \right)
\end{equation*}
integrates both local neighborhood features and the global query context, with \( \mathbf{e}_{q_r}^\ell \) being the query-specific relation embedding refined by the LSTM module.

\textbf{Biological Relevance Filtering} 
Following AdaProp~\citep{adaprop}, after node representations are updated, we compute an importance score for each candidate node \( t \):
\begin{equation}
\mathbf{s}_t = \mathbf{W}_\text{samp} \cdot \mathbf{h}_t^\ell(q_e, q_r)
\end{equation}
This score quantifies the biological relevance of each node. We then filter the candidate set by retaining only the top \( K \) nodes. During training, the top \( K \) nodes are selected via a differentiable Gumbel-Softmax, while during inference, a conventional Softmax selection is applied:
\begin{equation}
\mathcal{V}^{(\ell)} = \text{TopK}\left(\mathbf{s}_t \mid t \in \mathcal{C}^{(\ell)}\right).
\end{equation}
For details on gradient-preserved hard selection, see SM2.

\textbf{Final Subgraph Construction} 
Building upon this iteration, the final subgraph \( \mathcal{G}_q \) is constructed over \( \ell \) layers (Fig.~\ref{fig:overall}~(D)):
\begin{equation}
\mathcal{G}_q = (\mathcal{V}_q, \mathcal{E}_q)
\end{equation}
where \( \mathcal{V}_q \) denotes the set of selected entities and \( \mathcal{E}_q \) the relationships among them. This refined subgraph, enriched with contextually relevant information, is then used for downstream tasks such as knowledge graph completion and reasoning, ensuring that only the most pertinent interactions are propagated.

\subsubsection{Joint Formulation of Scoring and Loss Functions}
\label{sec:score}
Focusing only on graph message or knowledge representation in the final scoring function may miss complementarity. Pure graph modeling may overlook deeper semantic relationships, whereas embedding methods might not capture fine-grained structural details.
To better leverage the advantages of both perspectives, BioGraphFusion incorporates elements from knowledge embedding and graph propagation into its final scoring function.
For a triple \((q_e, q_r, q_a)\), our score is defined as a weighted sum (see Fig.~\ref{fig:overall}~(E)):
\begin{equation}
\tilde{f}(q_e, q_r, q_a) = \lambda\, f(q_e, q_r, q_a) + (1-\lambda)\, \phi(q_e, q_r, q_a)
\label{score}
\end{equation}
where \(\lambda \in [0,1]\) balances the contributions from two key components. \(f(\cdot)\) represents the score derived from the KE-informed graph propagation process, capturing contextualized structural patterns, whereas \(\phi(\cdot)\) provides a direct global semantic score obtained through tensor decomposition.
The hybrid design combines semantic knowledge with graph propagation through bidirectional interactions to refine structural representations. 
The component \(f(q_e, q_r, q_a)\) is computed from the final representation of the target entity obtained via iterative message passing:
\begin{equation}
    f(q_e, q_r, q_a) = \mathbf{w}^\top \mathbf{h}_{q_a}^{\ell}(q_e, q_r)
\end{equation}
and the tensor decomposition–based score, capturing the global biological context, is given by
\begin{equation}
    \phi(q_e, q_r, q_a) = \sum_{d=1}^{D} \mathbf{e}_{q_e}^{(d)} \cdot \mathbf{e}_{q_r}^{(d,\ell)} \cdot \mathbf{e}_{q_a}^{(d)}
\end{equation}
with \(\mathbf{e}_{q_e}^{(d)}\), \(\mathbf{e}_{q_a}^{(d)}\), and \(\mathbf{e}_{q_r}^{(d,\ell)}\) denoting the \(d\)-th components of the CP embeddings for the query entity, target entity, and the refined query relation (updated at layer \(\ell\) using an LSTM that incorporates \(\mathbf{e}_{q_e}\)), respectively. 

To train BioGraphFusion for biomedical completion and reasoning, we design a composite loss function with two objectives: (i) to maximize the likelihood of true relationships and (ii) to learn robust, generalizable embeddings. The primary component is a multi-class log-loss that encourages the model to assign higher scores to positive triples from the training set \(\mathcal{F}_{\text{tra}}\) compared to negative candidates. 
Specifically, the log-loss is defined as:

\begin{equation*}
\mathcal{L}_{\text{log}} = \sum_{(q_e,q_r,q_a) \in \mathcal{F}_{\text{tra}}} \left[-\tilde{f}(q_e,q_r,q_a) + \log \sum_{t \in \mathcal{V}} \exp\bigl(\tilde{f}(q_e,q_r,t)\bigr)\right]
\end{equation*}

In addition, following CP-N3~\citep{cp-n3}, we incorporate an N3 regularization term. The primary motivation for this is to penalize large magnitudes in CP embeddings, thereby mitigating overfitting:
\begin{equation}
R_{\text{N3}} = | \mathbf{e}{q_e} |_3^3 + | \mathbf{e}{q_r}^{\ell} |_3^3 + | \mathbf{e}_{q_a} |_3^3
\end{equation}
To further validate our choice of N3, ablation studies on regularization were conducted (see SM8). These studies have confirmed the robustness of our model architecture, demonstrating that the model performs well and outperforms baselines even when employing naive regularizations (L1 or L2). Notably, the N3 regularization, generally yields superior results over alternatives. This advantage is attributed to the selection of the optimized configuration, reinforcing its suitability for our approach.

Thus, the overall loss is given by:
\begin{equation}
\mathcal{L} = \mathcal{L}_{\text{log}} + \gamma R_{\text{N3}}
\end{equation}
where \(\gamma\) controls the regularization strength. 

\section{Experiments and Results}

\begin{table*}[htbp]
\centering % Added for better centering if the table is slightly narrower than textwidth
 \caption{Evaluation Results of BioGraphFusion on Biomedical Completion and Reasoning.\label{result}}
 \begin{tabular*}{\textwidth}{@{\extracolsep{\fill}}l|l|cccccccccc@{\extracolsep{\fill}}}
\toprule
\multirow{2}{*}{\textbf{Type}} & \multirow{2}{*}{\textbf{Models}} & \multicolumn{3}{c}{Disease-Gene Prediction } & \multicolumn{3}{c}{Protein-Chemical Interaction } & \multicolumn{3}{c}{Medical Ontology Reasoning} \\
\cline{3-5}\cline{6-8}\cline{9-11}
& & \textbf{MRR} & \textbf{Hit@1} & \textbf{Hit@10} & \textbf{MRR} & \textbf{Hit@1} & \textbf{Hit@10} & \textbf{MRR} & \textbf{Hit@1} & \textbf{Hit@10} \\
\midrule
 \multirow{5}{*}{\textbf{KE}}

 & RotatE & 0.263 & 0.202 & 0.381 & 0.606 & 0.512 & 0.778 & 0.925 & 0.863 & 0.993 \\

 & ComplEx& 0.392 & 0.336 & 0.498 & 0.356 & 0.236 & 0.594 & 0.630 & 0.493 & 0.893 \\

 & DistMult& 0.258 & 0.198 & 0.375 & 0.120 & 0.045 & 0.276 & 0.569 & 0.461 & 0.797 \\

 & CP-N3& 0.207 & 0.151 & 0.312 & 0.089 & 0.029 & 0.189 & 0.300 & 0.134 & 0.750 \\

& KDGene & 0.384 & 0.321 & \underline{0.523} & 0.085 & 0.023 & 0.170 & 0.260 & 0.100 & 0.708 \\

\midrule

\multirow{5}{*}{\textbf{GNN}}

 & pLogicNet & 0.228 & 0.173 & 0.335 & 0.591 & 0.564 & 0.630 & 0.842 & 0.772 & 0.965 \\

& CompGCN& 0.252 & 0.191 & 0.367 & 0.614 & 0.576 & 0.676 & 0.907 & 0.867 & 0.994 \\

& DPMPN& 0.293 & 0.235 & 0.393 & 0.632 & 0.614 & 0.729 & 0.930 & 0.899 & 0.980 \\

& AdaProp& 0.345 & 0.296 & 0.438 & \underline{0.662} &\underline{0.631} & 0.781 & \underline{0.969} & \underline{0.956} & \textbf{0.995} \\
& RED-GNN& \underline{0.389} & \underline{0.332} & 0.468 & \underline{0.662} & 0.613 & \underline{0.782} & 0.964 & 0.946 & 0.990\\

\midrule

\multirow{4}{*}{\textbf{Ensemble}}

& KG-BERT& - &  - &  - &  - &  - &  - & 0.774 & 0.649 & 0.967 \\ % Placeholder - fill these in

 & StAR& 0.247 & 0.192 & 0.361 & 0.426 & 0.326 &0.700 & 0.834& 0.720 & 0.976 \\ %
 
 & LASS & 0.211 & 0.167 & 0.324 &  0.401 & 0.314 & 0.691 & 0.908 & 0.952 & 0.983 \\ % Placeholder - fill these in

 & \textbf{ours} & \textbf{0.429$^{\ast}$$^{\ast}$} & \textbf{0.377$^{\ast}$$^{\ast}$} & \textbf{0.529$^{\ast}$} & \textbf{0.702$^{\ast}$$^{\ast}$} & \textbf{0.657$^{\ast}$} & \textbf{0.795$^{\ast}$} & \textbf{0.974} & \textbf{0.963$^{\ast}$} & \underline{0.991}\\

\botrule

\end{tabular*}
 \begin{tablenotes}
        \item  
        '-' means unavailable results.
        The best results are highlighted in \textbf{bold} and the second-best results are \underline{underlined}.
        '$\ast$' denotes statistically improvements over the best baseline ($\ast$: $p$-value$<0.01$; $\ast$$\ast$: $p$-value$<0.001$; paired t-test on 5 random seeds).
 \end{tablenotes}
\vspace{-1em}
\end{table*}

\begin{figure*}[htbp]%调节图片位置，h：浮动；t：顶部；b:底部；p：当前位置 
\centering %{\color{black!20}\rule{213pt}{37pt}}
\includegraphics[width=.9\textwidth]{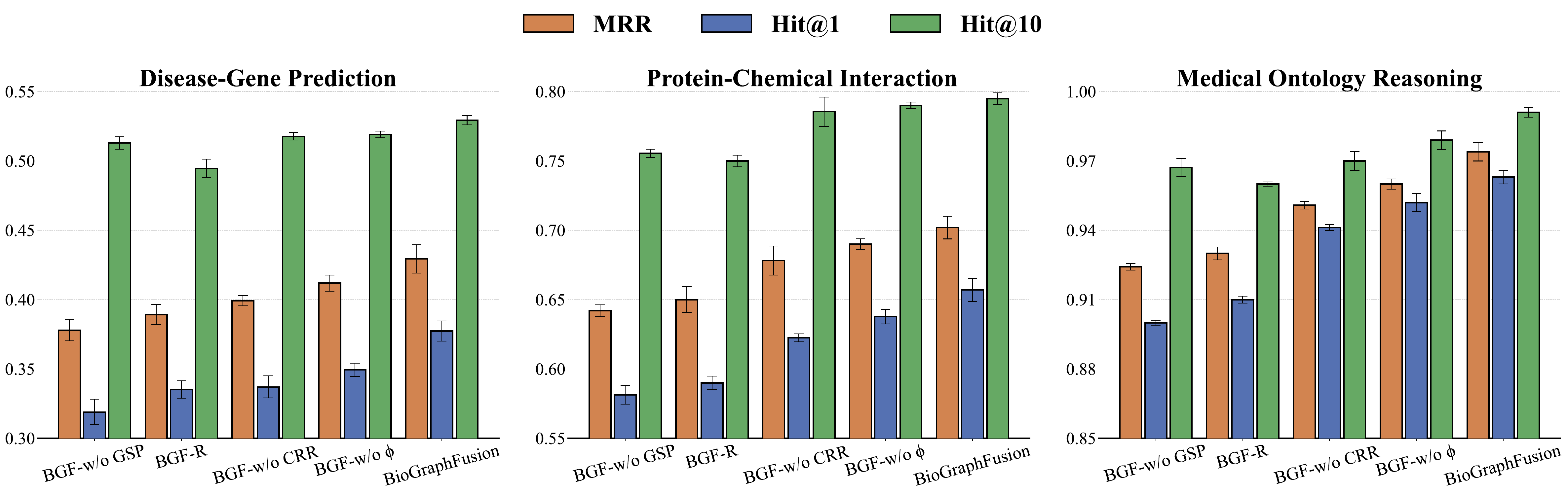}
\vspace{-1em}
\caption{Ablation study results for BioGraphFusion (BGF) across three biomedical reasoning tasks: disease-gene prediction, protein-chemical interaction, and medical ontology reasoning. Performance metrics include MRR, Hit@1, and Hit@10. The full model is compared against four ablated variants: BGF-w/o GSP, BGF-R, BGF-w/o CRR, and BGF-w/o~$\phi$.}\label{fig:ablation} 
\vspace{-1em}
\end{figure*}
% 注释图表
\subsection{Implementation Details}\label{sec:expdetails}

\textbf{Experimental Setup.}
All experiments were implemented in Python using PyTorch v1.12.1 and PyTorch Geometric v2.0.9 on a single NVIDIA RTX 3090 GPU. Key hyperparameters were tuned over specific ranges; detailed configurations are provided in SM3.\\
\textbf{Evaluation Metrics and Baseline Competitors.}
Following \citep{adaprop,redgnn}, we evaluate model performance using filtered ranking-based metrics: mean reciprocal rank (MRR) and Hit@\(k\) (with \(k=1\) and \(10\)). Detailed definitions of these metrics are provided in SM4. We benchmark BioGraphFusion against state-of-the-art methods from three major categories: Knowledge Embedding (KE) models, Graph Structure Propagation (GNN-based) approaches, and Ensemble methods. All baselines are implemented using publicly available code from the respective authors. Comprehensive descriptions of these baselines and implementation details are provided in SM5.\\
\textbf{Datasets and Data Integration.}
The Disease-Gene Prediction task uses 130,820 disease-gene associations from DisGeNET~\citep{DisGeNET}, partitioned 7:2:1 (training:validation:test) based on a specific fold from the KDGene~\citep{kdgene} 10-fold cross-validation setup. For comprehensive generalization assessment, we also conduct full 10-fold cross-validation (see SM6). Supplementary data include 14,631 drug-disease relationships from SIDER~\citep{sider} and 277,745 protein-chemical interactions from STITCH~\citep{stitch}. The Protein-Chemical Interaction task uses 23,074 interaction triples from STITCH, filtered to the top 100 most frequent genes~\citep{knowddi}, and partitioned 7:2:1. To address data imbalance from extensive background knowledge, we cap supplementary samples at 15,000 for Disease-Gene Association Prediction and 10,000 for Protein-Chemical Interaction. The Medical Ontology Reasoning task is based on the UMLS Terminology~\citep{UMLS}, pre-split into background, training, validation, and test sets as in prior work~\citep{adaprop,redgnn}. Further dataset and task details are in SM1.

\vspace{-1em}
\subsection{Overall Performance}

% As shown in Table~\ref{result}, BioGraphFusion consistently outperforms the baselines across all three tasks. Note the table is split into two parts, reflecting bipartite modeling paradigms: KE branch (top) and  GNN branch (bottom). While KE-based models generally focus on learning semantic representations of entities and relations, GNN-based methods rely more heavily on iterative message passing to capture local topological information. The results suggest that neither paradigm alone can fully address the complexity of biomedical knowledge capture. By integrating the complementary strengths of both branches, BioGraphFusion surpasses all baselines in overall performance.

Table~\ref{result} shows BioGraphFusion consistently outperforms KE, GNN, and Ensemble baselines across all three tasks. Regarding computational efficiency, SM7 compares the inference time and MRR performance of BioGraphFusion with competitive baseline models on the UMLS dataset, analyzing the trade-off between their predictive performance and computational efficiency.

\textbf{Knowledge Embedding Methods} reveal limitations in pure embedding approaches. ComplEx achieves moderate success in Disease-Gene Association Prediction (MRR 0.392) by modeling asymmetric relations, while RotatE (MRR 0.263) struggles despite its sophisticated rotation-based relation modeling. CP-N3's poor performance in Protein-Chemical Interaction is more telling. While CP-N3 uses tensor decomposition, a principle foundational to our approach, its standalone application, lacking crucial integration with structural learning, highlights the limitations of relying solely on this embedding technique. Even KDGene, engineered for disease-gene associations using interactional tensor decomposition, achieves only 0.384 MRR, showing semantic modeling alone, without adaptive structural guidance, cannot fully capture intricate biomedical dependencies. 
% Across tasks, KE methods capture direct semantic associations but fail to leverage topological context.

\textbf{Graph Neural Network Approaches} show different strengths and limitations. RED-GNN performs strongly in Disease-Gene Prediction (MRR 0.389), and AdaProp excels in Protein-Chemical Interaction (MRR 0.662). However, their weakness of reliance on structural patterns is apparent when compared to BioGraphFusion, which demonstrates consistent improvements in these tasks. While AdaProp has a slight edge in highly structured UMLS tasks, the merit diminishes in more semantically complex biomedical scenarios. While effective for local connectivity, pure structural propagation lacks the global semantic context needed to interpret biological relationships.

\textbf{Ensemble Methods} exhibit limitations in biomedical contexts. KG-BERT performs moderately in medical ontology reasoning (MRR 0.774), while StAR and LASS show limited effectiveness in Disease-Gene Association Prediction. A key constraint is their textual encoding components' limitation by sparse entity information—biomedical entities are often identifiers or technical terms, not descriptive text. This yields shallow semantic embeddings, hindering effective structural integration. While these methods try to bridge semantic understanding with structural patterns (StAR via Siamese encoding, LASS via joint fine-tuning), their limitations show effective biomedical ensemble integration needs more than combining components.
% ; it demands adaptive interplay effective with limited textual data while maximizing structural patterns.

\textbf{BioGraphFusion's Superior Performance} stems from its innovative deep coupling between semantic understanding and structural learning. Unlike existing methods that combine these paradigms statically, our model enables dynamic co-evolution where semantic insights guide structural reasoning while structural discoveries enrich semantic understanding. This deep coupling effectively models the intricate, context-dependent relationships in biomedical knowledge graphs, resulting in significant performance improvements across diverse tasks. 

\subsection{Ablation Study}\label{sec:ablation}
% To validate the contribution of the KE components, we conducted a series of ablation studies by modifying key modules involved in global semantic modeling and hybrid scoring. Four variants are tested: (i) \textcolor{red}{Removal of GSP module (BGF-w/o GSP)}; (ii) Random Query Encoding (BGF-R), in which the CP-derived query embeddings are replaced with randomly initialized vectors, disrupting semantic alignment; (iii) Removal of Contextual Relation Refinement (BGF-w/o CRR), which omits the LSTM-based updates for relation embeddings; and (iv) Elimination of the Tensor Decomposition Score (BGF-w/o~\(\phi\)), which excludes the CP-based branch from the hybrid scoring function, relying solely on graph-derived structural signals.
To evaluate individual component contributions in BioGraphFusion, we performed ablation studies on key modules for global semantics, graph structure propagation, and their hybrid scoring. Four targeted variants were implemented: 
(i) Removal of Graph Structure Propagation (BGF-w/o GSP): removes dynamic structural learning to assess GSP's role in our model; (ii) Random Query Encoding (BGF-R), in which the CP-derived query embeddings are replaced with randomly initialized vectors, disrupting semantic alignment; (iii) Removal of Contextual Relation Refinement (BGF-w/o CRR), which omits the LSTM-based updates for relation embeddings; and (iv) Elimination of the Tensor Decomposition Score (BGF-w/o~\(\phi\)), which excludes the CP-based branch from the hybrid scoring function, 
% relying solely on graph-derived structural signals.
leaving only the contextualized structural patterns to drive the scoring mechanism.

% \textcolor{red}{
% To complement the analysis of these key modules, further ablation studies on regularization (detailed in SM8) confirmed BioGraphFusion's architectural robustness: the model performed well and outperformed baselines even with L1 or L2 regularization. The N3 regularization, used in our final model, generally yielded superior results over these alternatives, contributing to its optimized configuration.
% }

% \vspace{-1em}
\subsubsection{Performance Comparison}
% 注释图表
% Fig.~\ref{fig:ablation} presents the MRR, Hit@1, and Hit@10 scores for the full BioGraphFusion model and each ablation variant across the three tasks. The full model consistently outperforms all variants. In particular, both BGF-R and BGF-Z exhibit significant performance degradation, underscoring the importance of CP-based query initialization for maintaining semantic alignment with the global knowledge captured in the tensor. Similarly, the removal of LSTM-based contextual relation refinement (BGF-w/o CRR) leads to a performance drop,  particularly on complex relational reasoning tasks like Medical Ontology Reasoning, while excluding the tensor decomposition score (BGF-w/o~\(\phi\)) also reduces overall effectiveness. These results confirm that the KE components, namely tensor decomposition and contextual relation refinement—are not merely supplementary but fundamental to capturing the nuanced relationships necessary for accurate biomedical knowledge graph completion and reasoning.

Fig.~\ref{fig:ablation} summarizes the ablation study results, highlighting the distinct contributions of structural propagation (GSP) and knowledge embedding (KE) components. Removing the GSP module (BGF-w/o GSP) leads to the most pronounced performance drop across all tasks, underscoring the essential role of dynamic structural learning in capturing topological dependencies and facilitating effective knowledge integration. This result demonstrates that structural propagation is indispensable for modeling complex biomedical relationships that rely on multi-hop and context-dependent interactions.

In contrast, the other three ablation variants—random query encoding (BGF-R), removal of contextual relation refinement (BGF-w/o CRR), and elimination of the tensor decomposition score (BGF-w/o~\(\phi\))—primarily target KE-related modules. Each of these modifications results in significant but distinct performance declines. BGF-R confirms the necessity of CP-based semantic initialization for maintaining meaningful entity representations; BGF-w/o CRR highlights the importance of LSTM-driven contextual refinement for relation embeddings; and BGF-w/o~\(\phi\) demonstrates that optimal performance requires balancing global semantic signals with graph-derived structural patterns. Collectively, these findings confirm our model's success stems from the synergy between structural propagation and semantic embedding, not from either component alone.

\begin{figure*}[htbp] \centering \includegraphics[width=.99\textwidth]{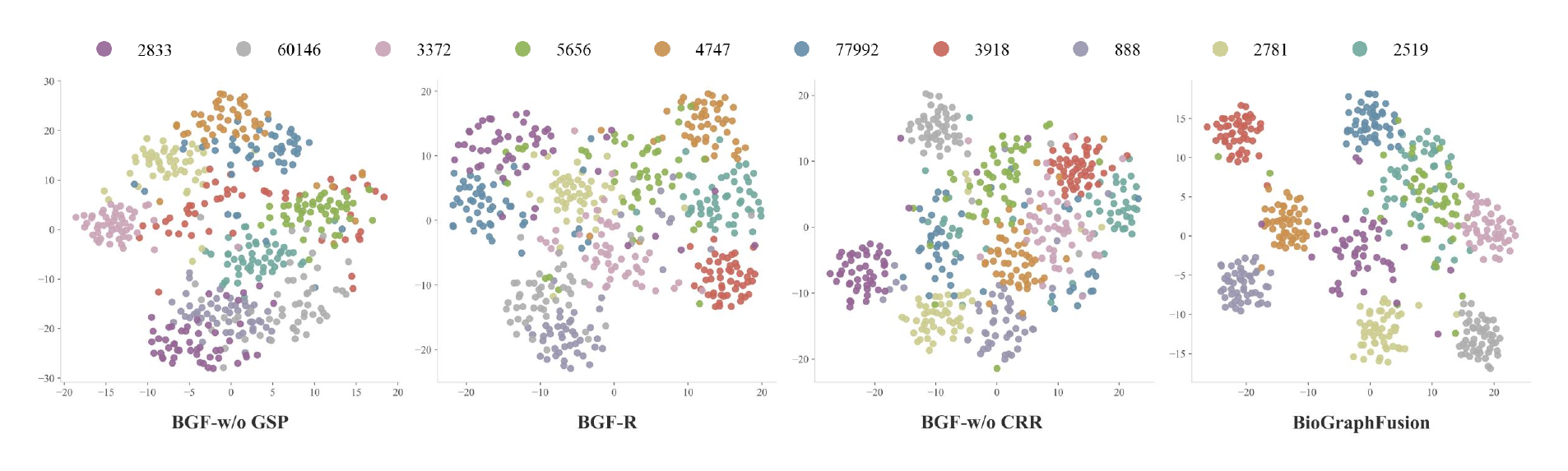} 
\vspace{-1em}\caption{t-SNE visualization of protein embeddings. Each subfigure shares the same proteins and each color represents proteins interacting with the same chemical compound, labeled by PubChem CID.}\label{fig:t-SNE} 
\vspace{-2em}
\end{figure*}

% \vspace{-1em}
\subsubsection{Semantic Embedding Visualization}
To assess KE components' impact on semantic representation, we visualize protein embeddings from the Protein-Chemical Interaction task using t-SNE (Fig.~\ref{fig:t-SNE}). We selected 10 chemical compounds (each linked to 50–100 proteins) and compared the full BioGraphFusion model with ablation variants BGF-w/o GSP, BGF-R, and BGF-w/o CRR. The BGF-w/o~\(\phi\) variant was excluded as its scoring function primarily affects prediction scores, not embedding coordinates. Protein embeddings were obtained via post-propagation representations for GSP variants (full, BGF-R, BGF-w/o CRR), and via final CP embeddings for BGF-w/o GSP.

The t-SNE visualizations in Fig.~\ref{fig:t-SNE} illustrate progressive improvement in semantic coherence as key architectural components are integrated. The full BioGraphFusion model produces optimally tight and well-separated protein embeddings for each chemical compound, showing strong intra-cluster cohesion and inter-cluster separation. Conversely, BGF-w/o GSP (relying solely on initial CP embeddings) shows the most diffuse clustering with indistinct inter-group boundaries, highlighting GSP's role in refining entity distinctions. BGF-R (with random query embeddings) exhibits clustering with significant overlap, confirming that effective GSP depends on high-quality initial semantic representations. BGF-w/o CRR shows clearer clustering than the previous two variants (benefiting from CP initialization and GSP), yet its clusters are less separated than the full model, emphasizing the crucial role of LSTM-driven relation refinement in forming clear, coherent clusters. 
These results confirm that CP initialization, dynamic GSP, and LSTM relation refinement each make unique contributions to meaningful biomedical entity representations. 
% These results show that BioGraphFusion is a unified framework that effectively produces meaningful biomedical entity representations through its integrated design.
Visualization results for competitive baselines in SM9 generally show more diffuse embedding clusters, further demonstrating BioGraphFusion's effectiveness.

\subsection{Hyperparameter Sensitivity Analysis}
We conducted extensive hyperparameter tuning on the disease-gene prediction task to examine the impact of key parameters on the final performance of BioGraphFusion. In our experiments, we varied the batch size, embedding dimension $D$, fusion weight $\lambda$, and the number of propagation steps $\ell$. Our results on the disease-gene dataset indicate optimal performance with a batch size of 16, an embedding dimension $D=32$, a fusion weight $\lambda=0.7$, and $\ell=6$ propagation steps.
Notably, the model enjoys robustness to batch size variations, while an embedding dimension of $D=32$ is found to effectively capture semantic details without over-parameterization.
Tuning $\lambda$ and $\ell$ reveals critical balances: $\lambda=0.7$ optimally harmonizes structural propagation with global semantic embeddings, while $\ell=6$ propagation step effectively balances information aggregation against over-smoothing.
This careful hyperparameter calibration is vital for maximizing model performance on biomedical tasks. Further details are in SM10.

\vspace{-1em}
\subsection{Case Analysis of CMM1}

%单张图片的插入格式
\begin{figure*}[htbp]%调节图片位置，h：浮动；t：顶部；b:底部；p：当前位置
\centering
% {\color{black!20}\rule{213pt}{37pt}}
 \includegraphics[width=.99\textwidth]{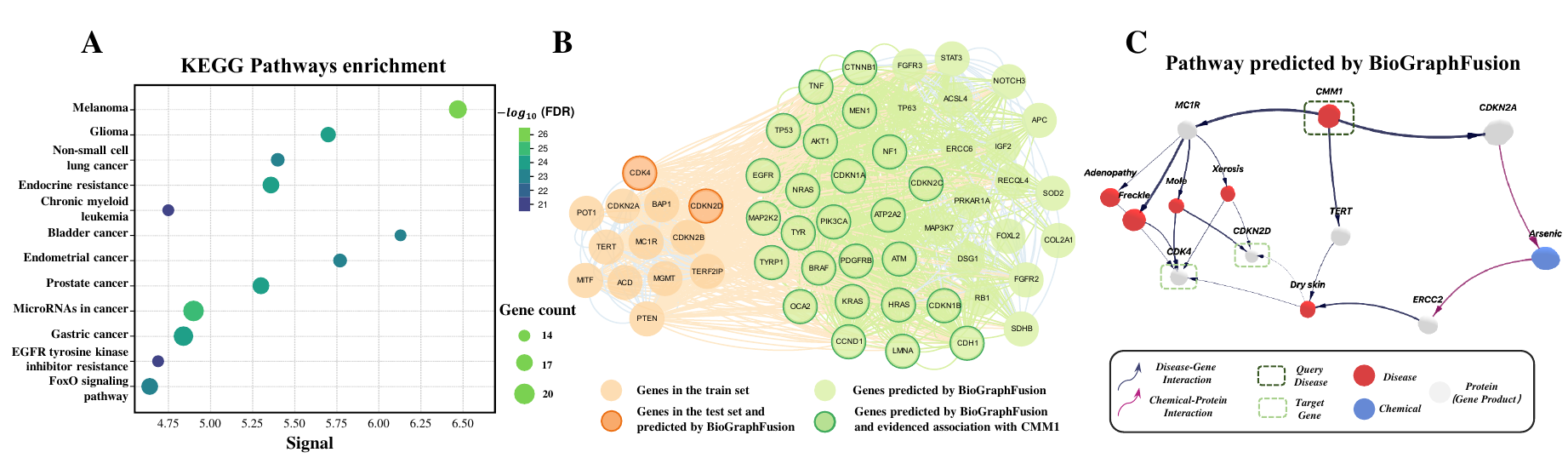}
 % \vspace{-1em}
\caption{Case study. \textbf{(A)} Analysis of KEGG Pathway Enrichment for the benchmark. The bubble chart shows significantly enriched pathways related to melanoma pathways.\textbf{(B)} Link visualization of known and predicted genes for melanoma on the PPI network.\textbf{(C)} Pathway predicted by BioGraphFusion from query disease CMM1 to melanoma-associated genes reveals a biologically plausible mechanistic link between CMM1 and established melanoma genes.}\label{fig:case} 
\vspace{-1em}
\end{figure*}

% \begin{table*}[htbp]
% \centering
% \caption{For CMM1, top 10 candidate gene predicted by BioGraphFusion. }
% \label{predict}
% \begin{tabular*}{\textwidth}{@{\extracolsep{\fill}}cccccc@{\extracolsep{\fill}}}
% \toprule
% \textbf{Rank} & \textbf{Predicted Gene} & \textbf{In Test Set?} & \textbf{PubMed Co-occ. IDs} & \textbf{In MalaCards} & \textbf{In ClinVar} \\ 
% \hline
% 1 & CDKN2D  & Yes &    -   &- & - \\ 
% 2  & CDK4    & Yes &   -    & -&-  \\ 
% 3  & AKT1   & No  &  38275910,39659584       &\checkmark & \checkmark  \\ 
% 4 & HPS1    & No &   15982315,23084991&  &  \\ 
% 5 & NF1    & No  &  38179395,37965626        &\checkmark & \checkmark  \\ 
% 6 & OCA2     & No  &  37646013,37568588       &\checkmark &  \checkmark \\ 
% 7 & TP53    & No  & 24919155,38667459         &\checkmark & \checkmark  \\
% 8 &  TYRP1    & No  &37646013,37239381        &\checkmark & \checkmark  \\
% 9 & TYR     & No  &   19578364,18563784       &\checkmark & \checkmark  \\ 
% 10 & NRAS    & No  &  38275910,38183141        &\checkmark & \checkmark  \\ 
% \bottomrule \end{tabular*} \end{table*}

% To validate BioGraphFusion’s capability in both identifying pathogenic candidates and inferring biologically plausible mechanisms, we conducted a comprehensive case study on cutaneous malignant melanoma susceptibility (CMM1). This analysis integrates gene prediction with multi-hop pathway reasoning, demonstrating how the model bridges genetic associations and molecular interactions to uncover context-specific disease mechanisms.

\subsubsection{Pathogenic Gene Prediction}

We used BioGraphFusion to predict ten candidate genes for CMM1, including two known disease-associated genes (CDKN2D and CDK4) and eight novel candidates (Table~\ref{predict}). To validate these predictions (Pred.), we cross-referenced the candidates against three independent databases: PubMed (using PMIDs), MalaCards and ClinVar. We found that seven of the eight novel candidate genes—AKT1 (rank 3), NF1 (rank 5), OCA2 (rank 6), TP53 (rank 7), TYRP1 (rank 8), TYR (rank 9), and NRAS (rank 10)—are documented in both MalaCards and ClinVar, indicating established associations with melanoma or related conditions. Additionally, PubMed searches revealed literature support for the co-occurrence of CMM1 with all eight novel candidates.
% BioGraphFusion also accurately predicted the known genes CDKN2D and CDK4, with top-ranked candidates such as AKT1 and TP53 aligning with prior reports~\citep{cho2015akt1, weiss1995mutation}.
\vspace{-1em}
\begin{table}[htbp]
\centering
\caption{For CMM1, top 10 candidate gene predicted by BioGraphFusion. }
\vspace{-1.5em}%%%%%%%%%%%%%%%%%%%%缩减竖直距离%%%%%%%%%%%%%%%%%%%%%%
\label{predict}
\begin{tabular*}{.49\textwidth}{@{\extracolsep{\fill}}ccccc@{\extracolsep{\fill}}}
\toprule
\textbf{Rank} & \textbf{Pred.} & \textbf{PMIDs} & \textbf{MalaCards} & \textbf{ClinVar} \\ 
\hline
1 & $^{*}$CDKN2D  &    -   &- & - \\ 
2  & $^{*}$CDK4D &   -    & -&-  \\ 
3  & AKT1  &  38275910,39659584       &\checkmark & \checkmark  \\ 
4 & HPS1   &   15982315,23084991&  &  \\ 
5 & NF1    &  38179395,37965626        &\checkmark & \checkmark  \\ 
6 & OCA2      &  37646013,37568588       &\checkmark &  \checkmark \\ 
7 & TP53     & 24919155,38667459         &\checkmark & \checkmark  \\
8 &  TYRP1      &37646013,37239381        &\checkmark & \checkmark  \\
9 & TYR     &   19578364,18563784       &\checkmark & \checkmark  \\ 
10 & NRAS     &  38275910,38183141        &\checkmark & \checkmark  \\ 
\bottomrule 
\end{tabular*} 
 \begin{tablenotes}
        \item *These genes predicted by BioGraphFusion are in the test set
 \end{tablenotes}
 \vspace{-2em}%%%%%%%%%%%%%%%%%%%%缩减竖直距离%%%%%%%%%%%%%%%%%%%%%%
 \end{table}
 % \vspace{-3em}

% \vspace{-1.2em}
\subsubsection{Pathway Enrichment and PPI Analysis}
To evaluate the biological relevance of both known genes and the predictions for CMM1, we performed a KEGG pathway enrichment analysis.
Fig.~\ref{fig:case}(A) presents the top 12 enriched pathways; notably, the “Melanoma” pathway shows the strongest enrichment (FDR = 2.20e-26) with 18 prominently represented genes. In addition, pathways associated with Glioma and Non-small cell lung cancer were also enriched, further supporting the biological plausibility of the candidate genes.
% Fig.~\ref{fig:case}(A) shows the top 12 enriched pathways, with the “Melanoma” pathway exhibiting the strongest enrichment (FDR = 2.20e-26) and including 18 prominently represented genes. Pathways related to Glioma and Non-small cell lung cancer were also significantly enriched, further supporting the biological plausibility of the candidate genes.

We further employed CMM1 as an illustrative example to evaluate the network proximity and functional coherence between genes in the train set and the candidate genes predicted by BioGraphFusion. For this analysis, we retain all 11 genes from the training set and 2 genes from the testing set of the DisGeNET dataset, and extract the top 50 candidate genes predicted by BioGraphFusion. The resulting Protein-Protein Interaction (PPI) network (Fig.~\ref{fig:case}(B)) exhibits markedly denser connectivity than would be expected by chance (P = 4.669E-86, binomial test). Detailed connectivity statistics and analysis are provided in SM11. This dense interconnectivity suggests that the candidates are functionally related to the known genes, reinforcing the biological relevance of our predictions.

\subsubsection{Pathway Reasoning and Biological Validation}\label{path}
By analyzing inference pathways that connect candidate genes to known disease ones, we aim to infer their functional relationships to discover the causative mechanism. For example, analyzing pathways linking disease CMM1 to known melanoma-associated genes CDKN2D and CDK4 (Fig.~\ref{fig:case}~(C)), with edge thickness representing attention weights, revealed a key pathway (CMM1 → MC1R → Mole → CDK4/CDKN2D) offering novel insights into melanoma pathogenesis.

% \textcolor{red}{
% To further understand the mechanisms captured by our model, we examined the inferred CMM1 $\rightarrow$ MC1R $\rightarrow$ Mole pathway, which is strongly supported by existing biological evidence. Research by~\citep{su2023melanocortin} indicates a progressive increase in MC1R expression throughout melanoma development, from benign moles to metastatic melanoma. Separately, \citep{van2020role}identified the MC1R \textit{Val60Leu} variant as a significant predictor of high mole counts, validating the MC1R-Mole link. These findings collectively support the biological plausibility of this pathway, suggesting a coherent mechanism in melanoma pathogenesis.}

To further understand the mechanisms our model hold, we examined the inferred CMM1 → MC1R → Mole pathway, strongly backed by existing biological evidence. Research by~\citep{su2023melanocortin} shows a progressive increase in MC1R expression throughout melanoma development, from benign moles to metastatic melanoma. Separately, \citep{van2020role} identified the MC1R \textit{Val60Leu} variant as a significant predictor for high mole counts, confirming the MC1R-Mole link. Together, these findings support this pathway's biological plausibility, suggesting a coherent mechanism in melanoma pathogenesis.

Notably, an alternative pathway, CMM1 $\rightarrow$ MC1R $\rightarrow$ Freckle, also receives high attention weights. This aligns with~\citep{bastiaens2001melanocortin}, who linked MC1R variants to freckle formation, reinforcing its connection to CMM1. As illustrated in Fig.~\ref{fig:case}(C), other MC1R-associated conditions, many with dermatological manifestations, show varying correlations with melanoma. These identified pathways deepen our understanding of disease mechanisms and highlight potential research directions.

% We focus on pathway-based reasoning and biological validation of candidate genes by BioGraphFusion. Analyzing inference pathways connecting candidate to known disease genes aims to infer functional relationships and find causative mechanisms.
% \vspace{-1em}
\section{Conclusion}
In this work, we introduce BioGraphFusion, a novel framework synergistically integrating semantic understanding with structural learning for biomedical KGC and KGR. BioGraphFusion enhances the dynamic interplay between these paradigms by using CP decomposition to establish a global semantic context. Building upon this, an LSTM-driven mechanism guides structural learning by dynamically refining relational information and updating semantic understanding during graph propagation. This enables learning context-dependent relation semantics and capture long-range dependencies, moving beyond static interpretations. Complemented by query-guided subgraph construction and a hybrid scoring mechanism, BioGraphFusion fosters a deep, adaptive refinement cycle between structural learning and semantic comprehension.
Experimental results show BioGraphFusion consistently outperforms traditional KE models, GNN-based approaches, and ensemble methods across biomedical benchmarks. Its ability to generate comprehensive features through an effective synergy of semantic insights and structural learning establishes it as a powerful tool. Finally, as demonstrated in the CMM1 case study, its capacity to uncover biologically meaningful pathways highlights its potential for advancing biomedical research.

\section*{Funding}
This work was supported in part by the Key Program of Natural Science Foundation of Zhejiang Province under Grant LZ24F030012, and the National Natural Science Foundation of China under Grant 62276232.

% \vspace{-1em}
%USE THE BELOW OPTIONS IN CASE YOU NEED AUTHOR YEAR FORMAT.
\bibliographystyle{abbrvnat}
\bibliography{reference}

\newpage

% =========================
% 补充材料部分开始
% =========================

% 补充材料内容（不包含文档类定义和包声明）
% 这个文件将被主文档包含

\clearpage
\onecolumn
\vspace{0.25cm}
\begin{center}
{\fontfamily{ptm}\selectfont\Huge\textbf{Supplementary Materials}}
\end{center}
\vspace{0.01cm}

\section*{Overall of the BioGraphFusion}

\begin{figure*}[htbp]%调节图片位置，h：浮动；t：顶部；b:底部；p：当前位置
\centering
% {\color{black!20}\rule{213pt}{37pt}}
 \includegraphics[width=.99\textwidth]{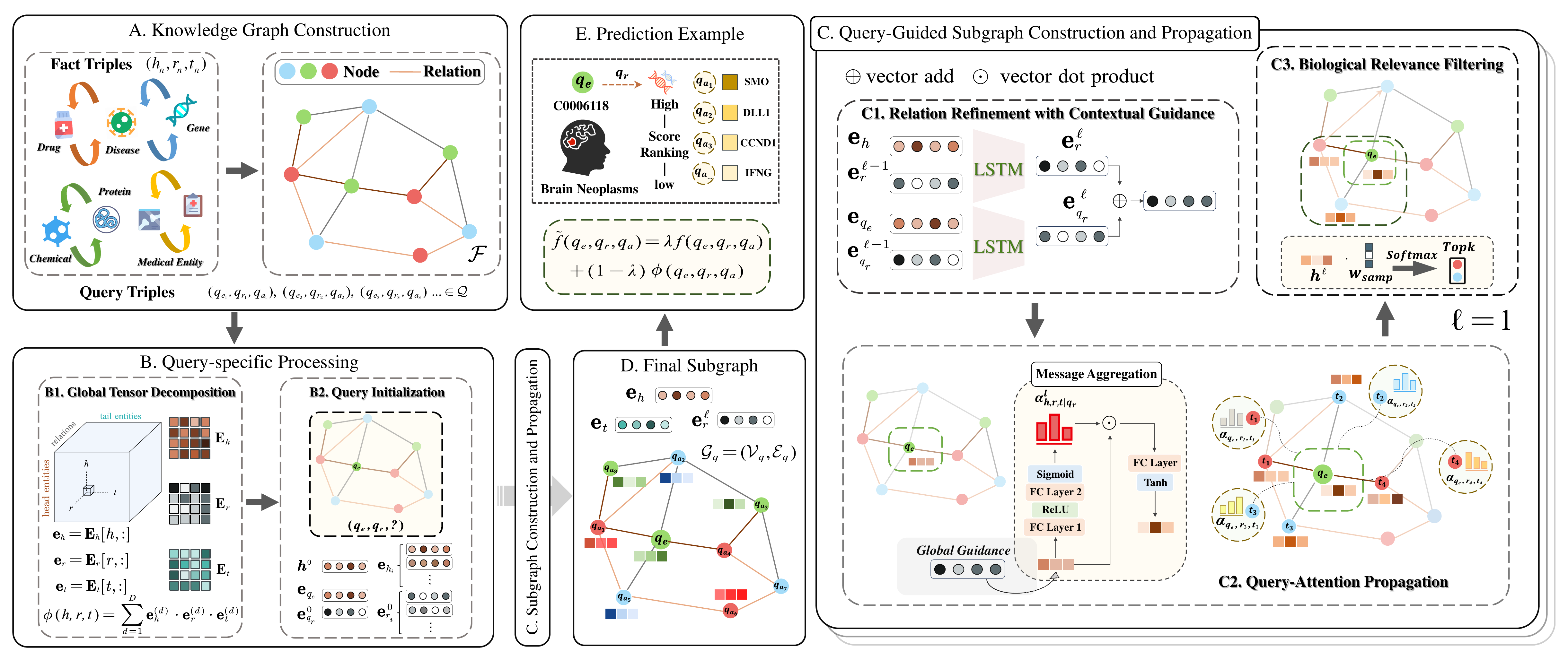}
\caption{Overview of the BioGraphFusion framework. (A) Knowledge Graph Construction: Integrating biomedical datasets to form a unified knowledge graph for downstream tasks. (B) Query-Specific Processing: A two-step process involving (B1) Global Tensor Decomposition that captures latent biological associations, and (B2) Query Initialization that guides the search process. (C) Subgraph Construction and Propagation, including (C1) Relation Refinement via LSTM, (C2) Query-Attention Propagation with context-based attention weights, and (C3) Biological Relevance Filtering to select the most pertinent entities. (D) Final Subgraph. (E) Scoring Integration that balances structural and semantic information and Prediction Example that selects the most promising predictions, with a focus on query disease Brain Neoplasms.}\label{fig:overall-new}
\end{figure*}

BioGraphFusion is a novel framework meticulously engineered to transcend the limitations of prior methods by enabling a truly synergistic integration of semantic understanding and structural learning within biomedical knowledge graphs (KGs). Its architecture, illustrated in Fig.~\ref{fig:overall-new}, is designed to foster deep interaction mechanisms, ensure dynamic structural learning under global semantic guidance, and achieve adaptive refinement through the coupled interplay of its components, thereby fostering a profound and dynamic coupling between these two paradigms.

The process commences with the construction of a biomedical KG from factual triples relevant to the specific task. For instance, Disease-Gene Prediction tasks might utilize graphs built from Drug-Disease and Protein-Chemical relationships, while Medical Ontology Reasoning would leverage diverse medical relationships to ensure domain-specific fidelity.

At the heart of BioGraphFusion's strategy to overcome the lack of overarching semantic direction in many previous models is its Global Biological Tensor Encoding module. This component employs Canonical Polyadic (CP) decomposition to generate initial, rich embedding matrices for head entities, relations, and tail entities. Crucially, these embeddings establish a foundational global semantic context, capturing latent biological associations and providing the essential top-down guidance for the subsequent, more nuanced and dynamic structural exploration that follows.

% 注释图表

Building upon this global semantic context, the Query-Guided Subgraph Construction and Propagation mechanism is introduced, playing a pivotal role in realizing BioGraphFusion's dynamic and adaptive capabilities. This module first iteratively constructs a query-relevant subgraph by selectively expanding along paths deemed semantically meaningful, guided by the global embeddings and the evolving query context. A core innovation within this propagation process is the LSTM-driven Contextual Relation Refinement. This powerful mechanism dynamically refines relation embeddings during multi-layer propagation within the focused subgraph. Such dynamic refinement allows these embeddings to adapt to evolving semantic contexts and capture long-range dependencies—a clear departure from static representations. Critically, this LSTM-driven process ensures that structural learning is continuously informed by and interacts with semantic insights, fostering a profound and dynamic coupling between these two paradigms. Complementing this deep, contextual refinement of relations, the query-guided nature of the subgraph construction further focuses structural exploration and message passing on the most biologically pertinent regions, enabling highly adaptive refinement and targeted learning.

Finally, to ensure a holistic and robust prediction, A hybrid scoring mechanism synthesizes the knowledge gained. This mechanism is aimed at achieving an optimal balance between the foundational global semantic knowledge (captured by the initial tensor embeddings) and the contextualized structural representations derived from the dynamic, semantically-guided graph propagation. This step ensures that both the broad, overarching associations and the fine-grained, dynamically uncovered structural patterns contribute to the final output.

In essence, by systematically establishing global semantic guidance, enabling dynamic and context-aware structural learning that is continuously informed by semantic insights, and ensuring their deep, adaptive coupling, BioGraphFusion's architecture is designed to deliver more accurate predictions and profound biological insights, moving beyond the capabilities of models with less integrated semantic and structural processing.

\section*{(1) SM1: Experimental Tasks and Datasets}
All experiments were conducted using the datasets presented in Table~\ref{dataset}. The table details three main tasks: prediction of disease genes, protein-chemical interactions, and medical ontology reasoning, each accompanied by specific background knowledge. Together, these datasets offer a comprehensive and robust foundation for evaluating our methods across multiple biomedical domains.

\begin{table}[htbp]
    \centering
    \caption{Datasets and Task Overview.}
    \label{dataset}
    \small
    \setlength{\tabcolsep}{1pt} % 控制表格列间距
    % 将 0.49\textwidth 修改为一个更大的值，例如 0.75\textwidth
    \begin{tabular*}{0.75\textwidth}{@{\extracolsep{\fill}}ccc}  % 根据需要调整列宽
        \toprule
        \multirow{2}{*}{ \textbf{Tasks} }               & \textbf{Background Knowledge}         &  \textbf{Main Datasets}         \\
                                                        & \textbf{Sources}                      &  \textbf{Targets}               \\
        \midrule
        \multirow{2}{*}{Disease-Gene }
        & Drug-Disease Relationships     \\
        \multirow{2}{*}{Prediction}
        &SIDER (14,631) & DisGeNet (130,820) \\
        & Protein-Chemical Relationships   &  Gene    \\
            &STITCH (277,745)  &                                                \\
        \midrule
        \multirow{2}{*}{Protein-Chemical}
        & Drug-Disease Relationships                 \\
        \multirow{2}{*}{Interaction }
        & SIDER (14,631)   &  STITCH (23,074) \\
        & Disease-Gene Relationships &  Chemical \\
           & DisGeNet (130,820)                              \\
        \midrule
        Medical Ontology
        & Various Medical Relationships          & UMLS(2,523)         \\
        Reasoning &  UMLS(4,006)   &  Multi-domain Entities \\
        \bottomrule % 原文是 \botrule，标准 booktabs 命令是 \bottomrule
    \end{tabular*}
\end{table}

\textbf{Disease-Gene Prediction task:}
This task aims to identify gene entities associated with a given disease; for example, if the input is "Alzheimer's Disease", the model would predict associated genes such as "APOE". We utilize 130,820 disease-gene associations from the DisGeNET database, following the data foundation of the KDGene study. To provide broader contextual information, this primary dataset is enriched by incorporating drug–disease relationships from SIDER~\citep{sider} and protein–chemical interactions from STITCH~\citep{stitch}. While these sources offer a substantial number of potential contextual interactions (14,631 and 277,745 respectively), to mitigate potential data imbalance from such extensive background knowledge and maintain a focused set of supplementary data, the number of supplementary samples utilized from these enrichment sources for the Disease-Gene Prediction task was capped at 15,000. Our primary experiments, detailed in this manuscript, were based on one specific fold selected from KDGene's 10-fold cross-validation setup (which uses a 90\% training/10\% testing split per fold). Within this chosen fold, KDGene's original 10\% segment served as our test set. The remaining 90\% (KDGene's training portion) was partitioned by us into a 70\% training set and a 20\% validation set for BioGraphFusion, relative to the total data in that fold. This established an effective 70\%/20\%/10\% train/validation/test configuration, where the validation set was used for hyperparameter optimization.

To provide a more comprehensive assessment of BioGraphFusion's generalization performance and robustness, and for direct comparison under KDGene's full benchmark protocol, we also conducted a complete 10-fold cross-validation. In this broader evaluation, for each of the 10 original KDGene folds, BioGraphFusion was trained directly on the entire 90\% training data portion defined by KDGene for that fold. For these 10-fold CV runs, we utilized the set of fixed hyperparameters (such as learning rate and embedding dimensions) that were determined as optimal during our initial single-fold experiments, and each fold was trained for a number of epochs consistent with those initial experiments. Subsequently, the model was evaluated on the corresponding 10\% original KDGene test set. These comprehensive 10-fold cross-validation results, alongside those for selected key baseline models under the same rigorous regimen, are detailed in the SM6: 10-fold Cross-Validation Results for Disease-Gene Prediction Task on DisGeNET.

\textbf{Protein–Chemical Interaction task:}
This task aims to predict chemical entities that interact with a given protein. We utilize 23,074 STITCH interaction triples filtered for the top 100 most frequent proteins~\citep{knowddi}, following the same 7:2:1 train-validation-test split. The model predicts chemical entities interacting with a given protein. For example, when presented with DRD2 (Dopamine Receptor D2), a key prediction would be acetylcholine (CIDm00000187). This specific interaction, annotated in the STITCH dataset as a protein-chemical relation, is biologically meaningful and reflects DRD2's established role in modulating cholinergic signaling. To address class imbalance from extensive negative sampling, supplementary samples are capped at 10,000 (disease-gene) and 10,000 (disease-drug).

\textbf{Medical Ontology Reasoning task:}
This task is a form of Knowledge Graph Reasoning (KGR) and utilizes the comprehensive UMLS Terminology~\citep{UMLS}. Following prior work~\citep{adaprop,redgnn}, this terminology is pre-split into background, training, validation, and test sets. Within this KGR framework, the task is specifically formulated as a link prediction problem: given a head concept and a specific UMLS relation type, the model must predict the correct tail concept to complete the triplet.
For instance, given the head UMLS concept Acquired\_Abnormality and the relation Result\_of, the model would aim to predict Phenomenon\_or\_Process as the tail; this specific prediction is ontologically justified because acquired medical conditions inherently result from various underlying phenomena or processes, thus forming the valid triplet (Acquired\_Abnormality, Result\_of, Phenomenon\_or\_Process). By concealing the tail entities of known relations during training and tasking the model with their prediction, reasoning is effectively transformed into a link prediction problem. 
This approach facilitates not only the completion of missing hierarchical links (e.g., predicting broader or narrower concepts using relations like “Isa” or “Part\_of”) and the discovery of implicit associations (e.g., through relations such as “Associated\_with” or “Co-occurs\_with”), but also the verification of consistency within medical domain knowledge.

\section*{(2) SM2: Gradient-Preserving Hard Selection Mechanism}

The Gradient-Preserving Hard Selection Mechanism in BioGraphFusion selects a fixed-size set of \( K \) most relevant entities (Top-K nodes) from a candidate set \( \mathcal{C}^{(\ell)} \) at each propagation layer \( \ell \). This focuses computational resources and directs information flow. A key challenge is performing this discrete Top-K selection differentiably for end-to-end training via gradient-based optimization.

The mechanism first computes continuous, soft relevance scores \( \mathbf{s}_t \) for each candidate node \( t \in \mathcal{C}^{(\ell)} \). During \textbf{training}, to achieve differentiable Top-K selection, BioGraphFusion employs the Gumbel-Softmax technique~\citep{GUMBEL,discrete}. This allows the derivation of a "hard" binary selection mask \( \mathbf{s}_t^\text{hard} \) (values near 0 or 1) from the soft scores \( \mathbf{s}_t \), indicating the selected Top-K nodes \( \mathcal{V}^{(\ell)} \):
\begin{equation}
\mathcal{V}^{(\ell)} = \text{TopK}\left(\mathbf{s}_t \mid t \in \mathcal{C}^{(\ell)}\right).
\end{equation}
The Gumbel-Softmax ensures this process is amenable to gradient-based learning.

To preserve gradients through this discrete selection when updating node representations \( \mathbf{h}_t^{(\ell)} \), the following strategy is used:
\[
\mathbf{h}_t^{(\ell)} \leftarrow \mathbf{h}_t^{(\ell)} \cdot \left(\mathbf{s}_t^\text{hard} - \text{detach}(\mathbf{s}_t) + \mathbf{s}_t\right)
\]
In the \textbf{Forward Pass}, this expression effectively multiplies \( \mathbf{h}_t^{(\ell)} \) by the hard selection mask \( \mathbf{s}_t^\text{hard} \). In the \textbf{Backward Pass}, the gradient with respect to the parameters generating \( \mathbf{s}_t \) is equivalent to the gradient of \( \mathbf{s}_t \) itself. This "straight-through estimator" variant allows discrete forward selection while using continuous soft scores for gradient estimation, preserving crucial gradient information.

During \textbf{inference}, differentiability is not needed, so a standard deterministic Top-K selection based on the highest soft scores \( \mathbf{s}_t \) is used.

This mechanism is vital as it enables focused, discrete node selection while maintaining end-to-end differentiability for effective training. It combines interpretable hard selections with stable gradient-based learning, enhancing BioGraphFusion's ability to capture salient information for accurate knowledge graph completion and reasoning.

\section*{(3) SM3: Experimental Configuration and Hyperparameter Optimization}
We implemented all experiments in Python using PyTorch v1.12.1 and PyTorch Geometric v2.0.9 on a single NVIDIA RTX 3090 GPU. Training time and GPU memory usage vary with the dataset and batch size. Hyperparameter tuning was performed over the following ranges (see Table~\ref{tab:Hyperparameter Settings}): learning rate from \(\{10^{-4}, 5 \times 10^{-4}, 10^{-3}, 5 \times 10^{-3}, 10^{-2}\}\); batch size from \(\{4, 8, 16, 32\}\); embedding dimension \(D\) from \(\{32, 48, 64, 80, 96\}\); selected entities count \(K\) from \(\{100, 300, 500, 800, 1000\}\); fusion weight \(\lambda\) from \(\{0.3, 0.4, 0.5, 0.6, 0.7, 0.8\}\); regularization coefficient \(\gamma\) from \(\{0, 0.001, 0.01, 0.1\}\); and propagation steps \(\ell\) from \(\{4, 5, 6, 7, 8\}\). Other hyperparameters were set following the AdaProp configuration~\citep{adaprop}. The optimal settings were selected based on the MRR metric evaluated on the validation set \(\mathcal{F}_{\text{val}}\), with training capped at 100 epochs.

\begin{table*}[htbp]
\centering
\caption{Hyperparameter Settings for BioGraphFusion Experiments}
\label{tab:Hyperparameter Settings}
% 列格式 @{\extracolsep{\fill}}cc 会在两列之间平均分配多余的宽度
\begin{tabular*}{0.61\linewidth}{@{\extracolsep{\fill}}cc}
\toprule
\textbf{Hyperparameters} & \textbf{Values}  \\
\midrule
Learning Rate  & \(\{10^{-4}, 5 \times 10^{-4}, 10^{-3}, 5 \times 10^{-3}, 10^{-2}\}\) \\
Batch Size  & \(\{4, 8, 16, 32\}\) \\
Embedding Dimension \(D\) & \(\{16, 32, 48, 64, 96\}\) \\
Selected Entities Count \(K\) & \(\{100, 300, 500, 800, 1000\}\) \\
Fusion Weight \(\lambda\) & \(\{0.3, 0.4, 0.5, 0.6, 0.7, 0.8\}\) \\
Regularization Coefficient \(\gamma\) & \(\{0, 0.001, 0.01, 0.1\}\) \\
Propagation Steps \(\ell\) & \(\{4, 5, 6, 7, 8\}\) \\
\bottomrule
\end{tabular*}
\end{table*}

The code implementation reflecting these optimal hyperparameter settings is publicly available in our open-source repository.

\section*{(4) SM4: Evaluation Metrics}
We evaluate our model's performance on the biomedical knowledge graph tasks using two standard ranking metrics: Mean Reciprocal Rank (MRR) and Hit@\(k\), following established practices~\citep{kdgene,redgnn,adaprop}. These metrics are chosen to quantify both the overall ranking quality and the model's precision in retrieving the correct entity within the top-\(k\) predictions.

The Mean Reciprocal Rank (MRR) is defined as:
\[
\text{MRR} = \frac{1}{|\mathcal{Q}|} \sum_{(q_e, q_r, q_a) \in \mathcal{Q}} \frac{1}{\text{rank}(q_e, q_r, q_a)}
\]
where \(\mathcal{Q}\) denotes the set of test queries in the form \((q_e, q_r, ?)\) and \(\text{rank}(q_e, q_r, q_a)\) indicates the position of the correct tail entity \(q_a\) for the query \((q_e, q_r, ?)\) in a filtered ranking list. In the filtered setting, all known true triples (except for the target triple) are removed from the candidate list to ensure that only plausible negatives are considered. MRR provides an average measure of ranking quality across all test queries, with higher values indicating better overall performance.

The Hit@\(k\) metric measures the proportion of queries for which the correct entity appears within the top \(k\) predictions. It is computed as:
\[
\text{Hit@}k = \frac{1}{|\mathcal{Q}|} \sum_{(q_e, q_r, q_a) \in \mathcal{Q}} \mathbb{I}\left(\text{rank}(q_e, q_r, q_a) \leq k\right)
\]
where \(\mathbb{I}(\cdot)\) is the indicator function that returns 1 if the condition is true and 0 otherwise. This metric reflects the model's precision in retrieving the correct answer among the top predictions.

\section*{(5) SM5: Baseline Models and Implementation Protocols}

\subsection*{Baseline Model Categories}
To evaluate the performance of our framework in biological knowledge graph analysis, we benchmark against state-of-the-art methods from three major categories: Knowledge Embedding (KE) models, Graph Neural Network (GNN-based) approaches, and Ensemble methods:

For KE, we benchmark against: RotatE~\citep{rotate} modeling relations as complex space rotations; ComplEx~\citep{complex} capturing asymmetric interactions via complex-valued embeddings; DistMult~\citep{dismult} using a simplified bilinear scoring function; CP-N3~\citep{cp-n3} which enhances the Canonical Polyadic (CP) decomposition with N3 regularization; and KDGene~\citep{kdgene}, a model specifically designed for disease-gene prediction using interactional tensor decomposition.

For GNNs, we compare with: pLogicNet~\citep{plogicnet} which integrates probabilistic logic with neural networks; CompGCN~\citep{compgcn} that incorporates relation composition into Graph Convolutional Networks; DPMPN~\citep{dpmpn} employing a dynamic programming message passing network; AdaProp~\citep{adaprop} which learns adaptive propagation patterns for multi-hop reasoning; and RED-GNN~\citep{redgnn}, a relational digraph-based GNN for knowledge graph reasoning.

For Ensemble methods, we include: KG-BERT~\citep{KG-BERT}, treating knowledge graph triples as textual sequences and fine-tuning pre-trained language models like BERT for triple plausibility scores; StAR~\citep{StAR}, a hybrid model augmenting textual encoding with graph embedding techniques, utilizes a Siamese-style encoder and learns representations by employing both a deterministic classifier for semantic plausibility and a spatial distance measurement for structural relationships; and LASS~\citep{LASS}, jointly embedding triplet natural language semantics and structure via pre-trained language model fine-tuning with a probabilistic loss.

\subsection*{Baseline Implementation and Evaluation Protocols}
We utilize publicly available code from the original authors whenever possible, with download links provided in Table \ref{baseline} to facilitate access to original implementations for further details and reproducibility.

To ensure a robust and fair comparison, we adopted a systematic approach to hyperparameter selection for all methods. For each baseline model, we began with the officially recommended or widely reported hyperparameter settings. We then carefully adjusted key parameters (such as batch size, learning rate, and training epochs) based on the characteristics of our biological knowledge graph datasets and preliminary results on a dedicated validation set, aiming to optimize each model’s performance. Importantly, the final hyperparameters for all models were selected according to their performance (e.g., MRR) on the same validation set. This consistent validation protocol ensures that all models were evaluated under comparable conditions, enabling a fair and meaningful performance comparison.

\begin{table*}[htpb]
  \centering
  \caption{Baselines with URLs to download the codes provided by the respective authors.}\label{baseline}
  \begin{tabular}{l|l}
    \toprule
    Baselines  & URLs \\
    \midrule
    RotatE~\citep{rotate} & \url{https://github.com/DeepGraphLearning/KnowledgeGraphEmbedding} \\
    ComplEx~\citep{complex} & \url{https://github.com/ttrouill/complex} \\
    DisMult~\citep{dismult} & \url{https://github.com/mana-ysh/knowledge-graph-embeddings} \\
    CP-N3~\citep{cp-n3} & \url{https://github.com/facebookresearch/kbc} \\
    KDGene~\citep{kdgene} & \url{https://github.com/sienna-wxy/KDGene} \\
    pLogicNet~\citep{plogicnet} & \url{https://github.com/DeepGraphLearning/pLogicNet} \\
    CompGCN~\citep{compgcn} & \url{https://github.com/malllabiisc/CompGCN} \\
    DPMPN~\citep{dpmpn} & \url{https://github.com/anonymousauthor123/DPMPN} \\
    AdaProp~\citep{adaprop} & \url{https://github.com/LARS-research/AdaProp} \\
    RED-GNN~\citep{redgnn} & \url{https://github.com/LARS-research/RED-GNN} \\
    KG-BERT~\citep{KG-BERT}&
    \url{https://github.com/yao8839836/kg-bert}\\
    LASS~\citep{LASS} &
    \url{https://github.com/jhshen95/LASS}\\
    StAR~\citep{StAR}&
    \url{https://github.com/wangbo9719/StAR_KGC}\\
    \bottomrule
  \end{tabular}
\end{table*}

\section*{(6) SM6: 10-fold Cross-Validation Results for Disease-Gene Prediction Task on DisGeNET}
To provide a fair and comprehensive evaluation of our BioGraphFusion model on the disease-gene prediction task, we employed a more rigorous 10-fold cross-validation experimental design. Since conducting complete 10-fold cross-validation for all baseline models would require substantial computational resources and time, we selected representative models that performed most competitively in each category from Table 1 in the main manuscript (KDGene representing the KE category, RED-GNN representing the GNN category, and StAR representing the ensemble learning category). We then ran experiments based on their publicly available code under the unified 10-fold cross-validation setting.

Table~\ref{tab:supp_10fold_cv_results_revised} presents the detailed performance of each model under 10-fold cross-validation, with results reported as ``Mean ± Standard Deviation" to provide a more comprehensive reflection of model performance and stability. As demonstrated in the table, BioGraphFusion consistently outperforms other baseline methods across all three key metrics: MRR (0.436±0.014), Hit@1 (0.382±0.007), and Hit@10 (0.537±0.020). The reasonable distribution of standard deviation values further reflects the model's stability and robustness across different data splits.

\begin{table*}[htbp!]
  \centering
  \caption{Cross-validation results for Disease-Gene Prediction on the DisGeNET dataset.
           All models were evaluated using the original 10-fold splits defined by KDGene~\citep{kdgene}.
           Results are reported as mean $\pm$ standard deviation across the 10 folds.}
  \label{tab:supp_10fold_cv_results_revised}
  \sisetup{
    round-mode=places,      % Round to specified precision
    round-precision=3,      % Number of decimal places to round to
    table-format=1.3,       % Format for the mean value: 1 digit before, 3 after decimal
    separate-uncertainty,   % Input uncertainty as X(Y) for X +/- Y
    table-auto-round=true,  % Automatically round numbers based on precision
    table-figures-uncertainty=3 % Number of figures for the uncertainty part
  }
  % Column order: Type (l), Model (l), MRR (S), Hit@1 (S), Hit@10 (S)
    \begin{tabular}{@{}l l S S S@{}}
      \toprule
      \textbf{Type} & \textbf{Model} & \multicolumn{1}{c}{\textbf{MRR}} & \multicolumn{1}{c}{\textbf{Hit@1}} & \multicolumn{1}{c}{\textbf{Hit@10}} \\
      \midrule
      KE & KDGene~\citep{kdgene}  & 0.378(016) & 0.315(009) & 0.518(017)  \\
      GNN & RED-GNN~\citep{redgnn} & 0.394(012) & 0.338(008) & 0.472(018) \\
      Ensemble & StAR~\citep{StAR}  & 0.241(015) & 0.185(010) & 0.354(021) \\
      \midrule % Added a midrule to separate baselines from the proposed model
      Proposed & BioGraphFusion (ours) & \bfseries 0.436(014) & \bfseries 0.382(007) & \bfseries 0.537(020) \\
      \bottomrule
    \end{tabular}
\end{table*}
  
Regarding standard deviation, we observe distinct patterns across different methods: KDGene shows a relatively higher standard deviation in MRR (0.016), which may reflect the sensitivity of knowledge embedding methods when handling different data splits; RED-GNN demonstrates a moderate level of standard deviation (e.g., 0.012 for MRR), indicating that graph structure-based methods maintain relative stability in local structure modeling; while StAR's higher standard deviation in Hit@10 (0.021) might stem from its ensemble approach responding differently to variations in data distribution. In comparison, BioGraphFusion maintains reasonable and relatively low standard deviations across most metrics (0.014 for MRR, 0.007 for Hit@1), particularly demonstrating excellent performance on the strict Hit@1 metric, which fully proves that our approach not only delivers superior performance but also exhibits good stability, capable of providing reliable predictions under various data conditions.

\section*{(7) SM7: Computational Efficiency Analysis and Predictive Accuracy on UMLS Dataset}
\begin{figure*}[htbp] \centering \includegraphics[width=.6\textwidth]{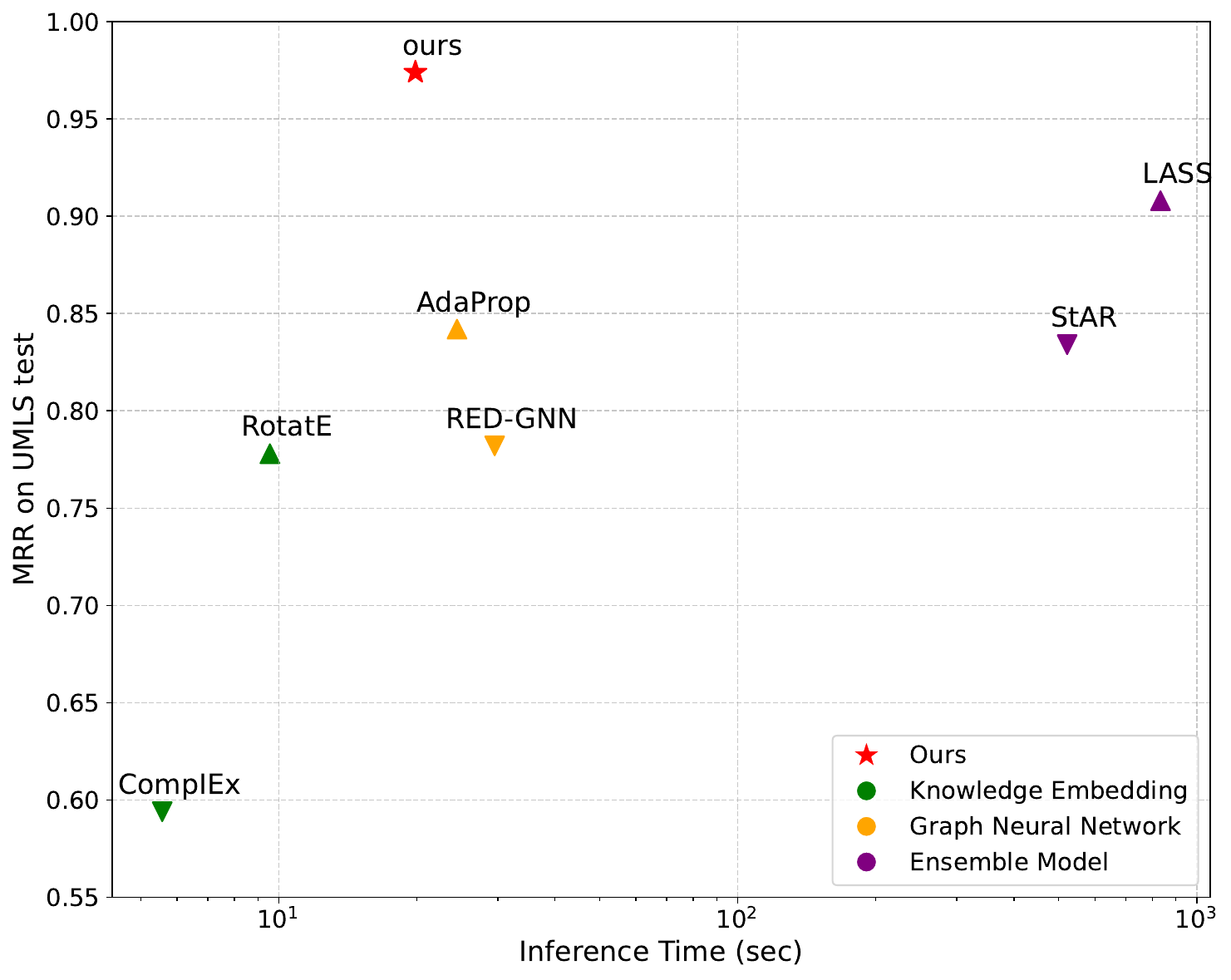} 
\vspace{-1em}\caption{Comparative analysis of computational efficiency and predictive accuracy on the UMLS dataset (inference time measured on NVIDIA RTX 3090 GPU). The scatter plot displays the Mean Reciprocal Rank (MRR) versus inference time (seconds) for BioGraphFusion (ours) and representative baseline models: Knowledge Embedding (RotatE, ComplEx), Graph Neural Network (AdaProp, RED-GNN), and Ensemble (LASS, StAR). Inference times are averaged over five runs using each model's optimal hyperparameters. This figure illustrates the balance between predictive performance and computational efficiency across different modeling approaches.}\label{fig:inference_mrr}
\vspace{-2em}
\end{figure*}
To provide a comprehensive assessment of the trade-off between computational efficiency and predictive accuracy, we present a comparative analysis of BioGraphFusion's inference performance relative to representative baseline models on the UMLS dataset. Inference times were averaged over five independent runs, all conducted under identical hardware and software conditions (single NVIDIA RTX 3090 GPU, PyTorch 1.12.1, Python 3.10.14).

For this comparative analysis, we selected high-performing and representative models from three main categories: Knowledge Embedding (KE) models (specifically RotatE and ComplEx), Graph Neural Networks (GNNs) (AdaProp and RED-GNN), and LM-based Ensemble approaches (LASS and StAR). Figure~\ref{fig:inference_mrr} provides a visual summary, plotting the Mean Reciprocal Rank (MRR) on the UMLS test set against the corresponding inference time (in seconds) for BioGraphFusion and these selected baselines. 

As depicted in Figure~\ref{fig:inference_mrr}, on the UMLS dataset, BioGraphFusion achieves a high MRR of 0.974 with an inference time of approximately 19.84 seconds. This positions BioGraphFusion effectively within the performance-efficiency landscape.
Its inference efficiency is notably strong, particularly when compared to the other GNNs evaluated; for instance, it is faster than both AdaProp (inference time: 24.45s, MRR: 0.842) and RED-GNN (inference time: 29.54s, MRR: 0.782), while also achieving a higher MRR. Relative to traditional KE methods, such as RotatE (inference time: 9.56s, MRR: 0.778) and ComplEx (inference time: 5.57s, MRR: 0.594), which are generally faster due to their simpler architectures, BioGraphFusion demonstrates a substantial improvement in MRR for a moderate increase in inference time.
Furthermore, BioGraphFusion is significantly more efficient in its inference phase than the typically more computationally demanding LM-based ensemble systems such as LASS (inference time: 833.89s, MRR: 0.908) and StAR (inference time: 522.0s, MRR: 0.834). BioGraphFusion not only boasts a much shorter inference time than these LM-based models but also achieves a higher MRR than both.

BioGraphFusion's favorable balance between high performance and efficient inference appears to stem from its sophisticated design that addresses key limitations of other paradigms. The framework's use of global semantic guidance, derived from tensor decomposition, likely enables more direct and interpretable pathfinding during reasoning. This global guidance can inform BioGraphFusion's query-guided subgraph construction, allowing it to selectively build more focused and semantically relevant subgraphs. This potentially streamlines the process compared to GNNs that might rely on more computationally intensive or exploratory subgraph generation strategies when lacking such strong initial global priors. Moreover, unlike some LM-based approaches that may not fully or efficiently integrate dynamic semantic insights with evolving structural graph updates during multi-hop reasoning, or traditional KE techniques that might oversimplify relational patterns and overlook critical graph structure, BioGraphFusion is designed to foster a deeper, adaptive interaction between semantic context and structural information. While some simpler KE methods offer faster raw inference speeds, BioGraphFusion strikes a compelling balance: achieving efficient processing within an expressive framework that supports complex, stable, and ultimately more effective knowledge graph reasoning by robustly modeling and leveraging the crucial interplay between semantics and structure. This demonstrates that BioGraphFusion can attain state-of-the-art predictive accuracy without incurring the prohibitive computational overheads observed in some other advanced, particularly LM-based, models.

\section*{(8) SM8: Regularization Studies and LSTM Validation}
\subsection*{Validation of LSTM for Contextual Relation Refinement}

To empirically validate our architectural choice of Long Short-Term Memory (LSTM) networks for the Contextual Relation Refinement Module, we conducted comprehensive comparative experiments where this critical component was systematically replaced with alternative neural architectures. As detailed in Table~\ref{tab:Variant_analysis}, the LSTM-based configuration of BioGraphFusion consistently demonstrated superior or highly competitive performance across all three evaluated datasets. For instance, on the Disease-Gene Prediction task, it achieved notable metrics (MRR 0.429, Hit@1 0.377, Hit@10 0.529), outperforming RNN and GRU variants. This trend of superior contextual modeling was similarly observed in the other biomedical tasks, reinforcing its general applicability.

This performance advantage directly stems from LSTM's architectural design, which proves exceptionally well-suited for biomedical knowledge graphs where relation semantics are highly entity-dependent. While RNN and GRU demonstrated diminished efficacy due to their simpler memory mechanisms, LSTM's sophisticated structure provides multiple critical advantages through its tripartite gating system and dedicated memory cell. The forget gate selectively discards irrelevant aspects of prior relation states, the input gate regulates incorporation of entity-derived information, and the output gate determines which aspects propagate as the new relation embedding. This architecture enables LSTM to effectively preserve long-term dependencies and perform fine-grained contextual adjustments that RNN and GRU cannot achieve with their simplified structures, particularly in complex biomedical contexts where relation meanings shift substantially based on specific entity pairs.

Even when compared to more complex architectures like Echo State Networks and Temporal Convolutional Networks, the LSTM-based configuration demonstrated superior performance. ESNs' static reservoir design inherently limits adaptive memory control compared to LSTM's flexible gating. Similarly, while TCNs excel at sequence-level pattern processing, they are less aligned with our task's demand for targeted entity-specific relation modulation. The Transformer architecture, while undeniably powerful for capturing global dependencies in many sequence processing tasks, showed inferior performance for our particular module focused on iterative, entity-specific refinement of individual relation embeddings when compared to our LSTM-based configuration.

LSTM's inherent support for stateful updates, where output depends on both previous state and current inputs, perfectly aligns with our goal of iteratively refining relation embeddings based on contextual entity information. This allows the model to learn complex, adaptive mappings that tailor relation representations to specific entity pairs—essential for modeling the intricate biological processes whose interpretation varies significantly with participating molecular entities.

These empirical findings and theoretical considerations robustly affirm LSTM as the optimal architecture for contextual relation refinement in BioGraphFusion, highlighting the importance of aligning neural architecture characteristics with specific task demands in complex biomedical domains.

\begin{table}[htbp]
  \centering
  \small % 减小字体大小以适应页面宽度
 \caption{Comparative Performance of BioGraphFusion Variants: Validating LSTM for Contextual Relation Refinement and Assessing the Impact of N3 Regularization. Best results are shown in \textbf{bold}, and second-best results are \underline{underlined}.}
  \label{tab:Variant_analysis}
  \begin{tabular*}{\textwidth}{@{\extracolsep{\fill}}lccccccccc@{}}
    \toprule
    \textbf{BioGraphFusion} & \multicolumn{3}{c}{\textbf{Disease-Gene}} & \multicolumn{3}{c}{\textbf{Protein-Chemical}} & \multicolumn{3}{c}{\textbf{Medical Ontology}} \\
    \textbf{Model variants} & \multicolumn{3}{c}{\textbf{Prediction}} & \multicolumn{3}{c}{\textbf{Interaction}} & \multicolumn{3}{c}{\textbf{Reasoning}} \\
    \cmidrule(lr){2-4} \cmidrule(lr){5-7} \cmidrule(lr){8-10}
     & \textbf{MRR} & \textbf{Hit@1} & \textbf{Hit@10} 
     & \textbf{MRR} & \textbf{Hit@1} & \textbf{Hit@10}
     & \textbf{MRR} & \textbf{Hit@1} & \textbf{Hit@10} \\
    \midrule
    w/ RNN   & 0.387 &  0.332 & 0.492 & 0.654 & 0.603 & 0.752 & 0.942 & 0.924 & 0.982   \\
    w/ GRU  & 0.386 & 0.330 & 0.495 & 0.659 & 0.608 & 0.755 & 0.951 & 0.936 & 0.984 \\
    w/ ESN & 0.407 & 0.344 & \underline{0.521} & 0.683 & 0.632 & 0.778 & 0.965 & 0.947 & \underline{0.990}  \\ 
    w/ TCN    & 0.404 & 0.353 & 0.515 & 0.689 & 0.639 & 0.782 & \underline{0.969} & \underline{0.952} & 0.988     \\
    w/ Transformer   &  0.383 & 0.321  & 0.500 & 0.671 & 0.621 & 0.765 & 0.953 & 0.934 & 0.985 \\
    \midrule
    w/ L1   & 0.409 & 0.358 & 0.518 & 0.687 & 0.637 & 0.780 & 0.961 & 0.945 & 0.987    \\
    w/ L2   & \underline{0.412} & \underline{0.364} & 0.520 & \underline{0.694} & \underline{0.645} & \underline{0.788} & 0.967 & 0.950 & 0.989 \\
    w/o N3   & 0.395 & 0.338 & 0.509 & 0.676 & 0.626 & 0.769 & 0.955 & 0.937 & 0.986  \\
    \midrule
    \textbf{Ours (w/ LSTM \& N3)} & \textbf{0.429} &
    \textbf{0.377} & \textbf{0.529} & \textbf{0.702} & \textbf{0.657} & \textbf{0.795} & \textbf{0.974} & \textbf{0.963} & \textbf{0.991} \\
    \bottomrule
  \end{tabular*}
\end{table}

 \subsection*{Analysis of N3 Regularization Impact}
This section assesses the role of N3 regularization within the BioGraphFusion framework, with a particular focus on determining whether it acts as a confounding factor for the model's performance improvements. N3 regularization is employed in BioGraphFusion, following established approaches such as CP-N3~\citep{cp-n3}, to operate on the CP embeddings, specifically \(\mathbf{e}_{q_e}\), \(\mathbf{e}_{q_r}^{\ell}\), and \(\mathbf{e}_{q_a}\). The technique targets the sum of cubes of embedding magnitudes. This form of regularization is intended to mitigate overfitting and enhance the model's ability to learn from complex biological knowledge graphs. We aim to demonstrate that BioGraphFusion's core architectural innovations provide substantial benefits independently, and that N3 regularization serves as a well-suited, complementary component rather than the sole driver of high performance.

To dissect the precise influence of N3 regularization, we compare our full BioGraphFusion model against several alternative configurations, as detailed in Table~\ref{tab:Variant_analysis}. These include variants of BioGraphFusion that incorporate standard L1 regularization (L1) and standard L2 regularization (L2), both applied to the same CP embeddings. Additionally, we evaluate a BioGraphFusion variant from which N3 regularization has been entirely removed. This comparative analysis allows for a clear delineation of N3's contribution relative to other regularization methods and its overall necessity.

This analysis first evaluated if BioGraphFusion's foundational architecture performs strongly without the specific N3 regularization. As shown in Table~\ref{tab:Variant_analysis}, on the Disease-Gene Prediction task, BioGraphFusion with L2 regularization (MRR 0.412) and L1 regularization (MRR 0.409) both surpassed the strongest competing baseline model. Similar strong performance with L1 and L2 regularization, relative to baselines, was observed across the other datasets as well. These results highlight that BioGraphFusion's primary efficacy stems from its core architectural innovations: the establishment of a global semantic foundation through Canonical Polyadic (CP) decomposition, dynamic structural reasoning actively guided by these initial embeddings, and advanced, context-aware relation refinement using LSTMs. This establishes that the model's competitive edge is rooted in this inherent design, which fosters a synergistic interplay between semantic understanding and structural learning, rather than depending solely on N3 regularization, thus confirming its robust foundational design.

While the core architecture is strong, regularization choice significantly impacts final performance. Our BioGraphFusion model with N3 regularization achieved the highest scores, notably outperforming its L2, L1, and non-regularized counterparts. This consistent performance hierarchy indicates that although any regularization is beneficial over none, N3 offers a distinct advantage over conventional L1 and L2 methods for this model.

The superior performance of N3 regularization suggests that its specific mechanism, which penalizes the ($l_3$)-norm of embedding magnitudes, is particularly well-aligned with the characteristics of the CP embeddings used and the demands of biological knowledge graph tasks. This type of regularization may offer a more suitable inductive bias for navigating the complex and often sparse relationships prevalent in such data. Consequently, it appears to lead to more effective prevention of overfitting and improved generalization capabilities compared to L1 or L2 norms in this specific context.

In essence, BioGraphFusion's primary effectiveness is rooted in its strong and innovative core architecture, which delivers competitive results independently. The N3 regularization then acts as a significant and synergistic enhancement, further elevating the model's performance to a state-of-the-art level.

\section*{(9) SM9: t-SNE Visualizations and Embedding Analysis with Baseline Models} % (假设您已采纳之前的标题建议)

To evaluate the semantic quality of embeddings learned by BioGraphFusion relative to other approaches, we conducted a comparative t-SNE visualization analysis (Fig.~\ref{fig:t-SNE}). This analysis focuses on protein embeddings from the Protein-Chemical Interaction task, specifically visualizing those associated with 10 chemical compounds (PubChem CIDs), each linked to 50–100 interacting proteins. BioGraphFusion's embeddings are compared against those from three representative baselines: RotatE (knowledge graph embedding)~\citep{rotate}, RED-GNN (GNN-based)~\citep{redgnn}, and StAR (ensemble text and structure)~\citep{StAR}.

\begin{figure*}[htbp] \centering \includegraphics[width=.8\textwidth]{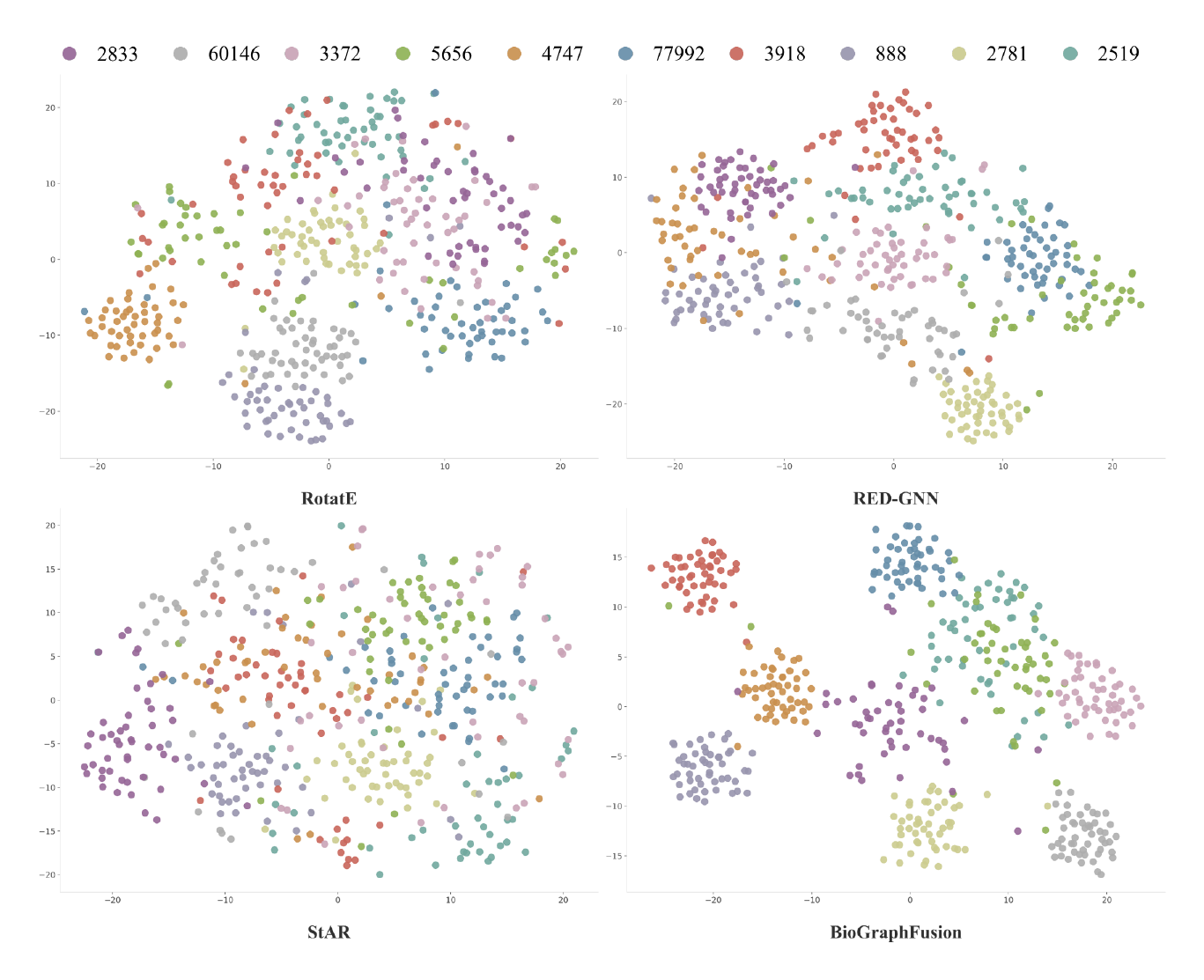} 
\vspace{-1em}\caption{t-SNE visualization of protein embeddings. Each subfigure shares the same proteins and each color represents proteins interacting with the same chemical compound, labeled by PubChem CID.}\label{fig:t-SNE} 
\vspace{-2em}
\end{figure*}

Protein embeddings for each method are obtained from their final learned representations. For GNN-based models like RED-GNN and our BioGraphFusion, these are typically generated after message propagation phases. For KE models like RotatE, embeddings are the direct output of the training process that optimizes a scoring function for triples. For StAR, embeddings are derived from its framework that integrates textual encoding with structural information. These embeddings are then reduced to two dimensions for visualization. The t-SNE plot corresponding to the BioGraphFusion model (shown in Fig.~\ref{fig:t-SNE}) yields tightly clustered and semantically coherent protein embeddings, indicating that the integration of CP decomposition-derived initial embeddings and LSTM-driven relation updates effectively shapes the representations to capture fine-grained interaction contexts.

In contrast, the selected baseline methods, as depicted in their respective t-SNE visualizations (Fig.~\ref{fig:t-SNE}), tend to exhibit more diffuse embedding clusters with notable inter-group overlap.
RotatE's visualization shows that while this method is effective at modeling relational patterns through rotations in complex space, its representations (learned primarily based on existing triple structures) may not always group entities optimally based on the specific, nuanced semantic context of diverse protein-chemical interactions if such distinct contexts are not explicitly and sufficiently captured by distinct relational paths. This can potentially lead to less distinct clusters for proteins interacting with different chemicals.
For RED-GNN, its strength lies in capturing structural information and multi-hop relations via its relational digraph structure and query-attentive message passing mechanisms. While proficient in reasoning over graph structures, its inherent semantic differentiation for entities primarily depends on the learned relation embeddings and topological neighborhood information. In our experimental setup, the input features to RED-GNN did not contain rich initial semantic information beyond identifiers, and its GNN architecture primarily focused on propagating structural signals. Consequently, this resulted in broader, less separated clusters in its t-SNE plot compared to models that more deeply integrate explicit semantic features or leverage descriptive inputs.

StAR's t-SNE visualization is also considered. This model is designed as a hybrid approach augmenting textual encoding of triples with graph embedding techniques, using a Siamese-style textual encoder. Its ability to form distinct semantic clusters is significantly influenced by the availability and richness of descriptive text for proteins and chemicals. In our experimental setup for the PCI task, to maintain consistency in the input features across all compared methods, detailed textual descriptions for entities were not incorporated for any model; entities were primarily represented by their identifiers or brief names. Consequently, StAR's powerful textual encoding capabilities, which rely on such rich descriptions, were not fully leveraged in this setting. The model, therefore, likely relied more heavily on its structural learning components or the default textual interpretation of these basic identifiers. This absence of detailed descriptive semantic input may have contributed to less defined separations between protein groups in the visualization, as the model might not have captured the subtle distinguishing features necessary for tight, well-separated clustering based on their specific chemical interactions.

\section*{(10) SM10: Hyperparameter Sensitivity Analysis on the Disease-Gene Prediction Task}  
\begin{figure*}[htbp]%调节图片位置，h：浮动；t：顶部；b:底部；p：当前位置
\centering
% {\color{black!20}\rule{213pt}{37pt}}
 \includegraphics[width=.7\textwidth]{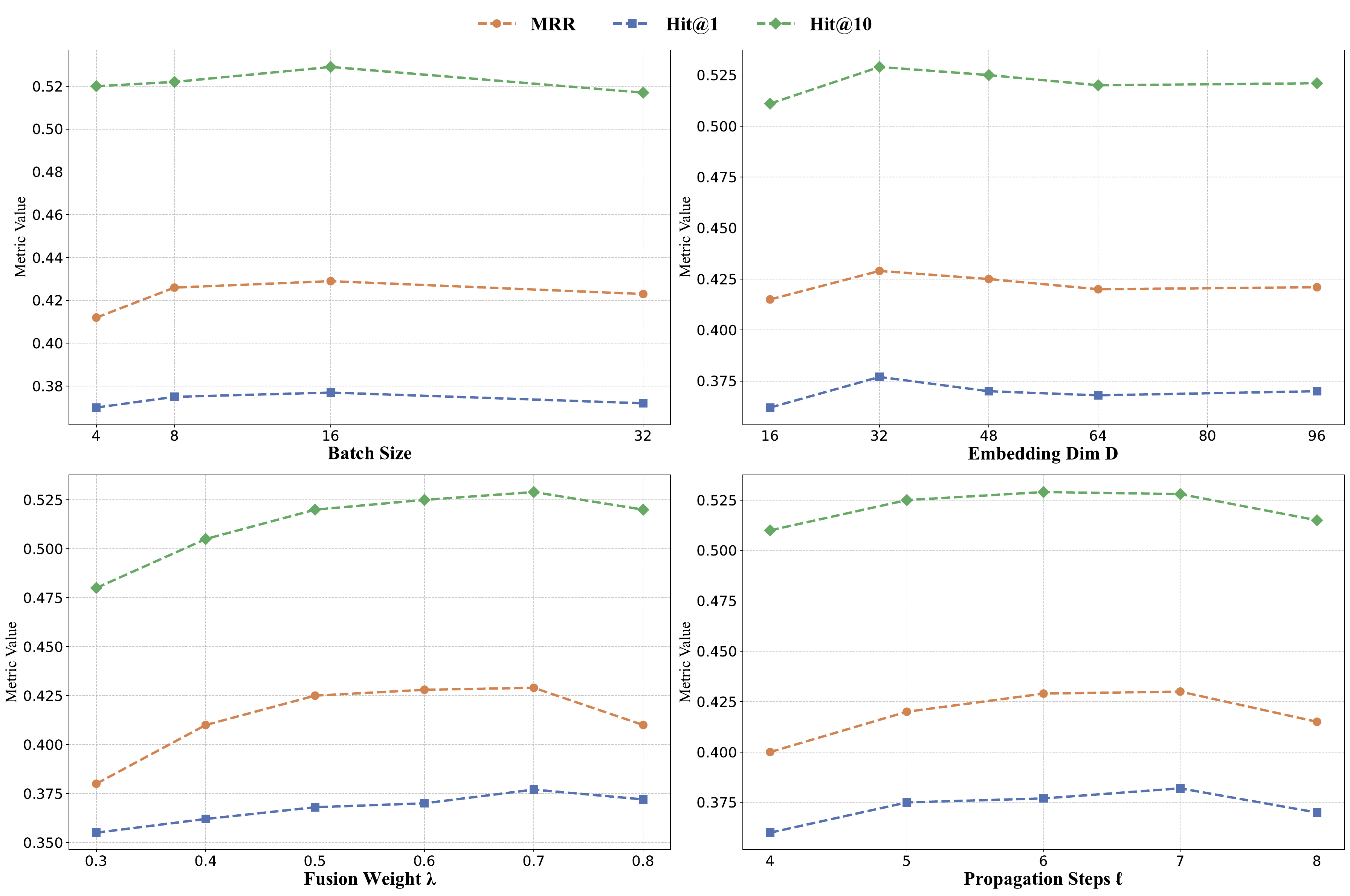}
\caption{BioGraphFusion underwent a hyperparameter sensitivity analysis to evaluate the influence of batch size, embedding dimension (\(D\)), fusion weight \(\lambda\), and propagation steps (\(\ell\)) on the resulting performance, as measured by MRR, Hit@1, and Hit@10.}\label{fig:parameter_sensitivity} 
\end{figure*}

To examine how key hyperparameters influence the predictive performance of BioGraphFusion, we conducted a sensitivity analysis on the disease-gene prediction task. We systematically varied batch size, embedding dimension \(D\), fusion weight \(\lambda\), and the number of propagation steps \(\ell\), observing their effects on ranking metrics such as MRR, Hit@1, and Hit@10. The analysis helps identify optimal parameter settings that maximize model effectiveness while maintaining stability across different configurations. The following sections detail the observed trends for each hyperparameter, with results visualized in Fig.~\ref{fig:parameter_sensitivity}.  
%单张图片的插入格式

\subsection*{Impact of Batch Size }
The selection of an appropriate batch size is crucial for balancing model performance and training efficiency. Our search space for this hyperparameter was guided by the common practice of using power-of-2 batch sizes for computational efficiency and the practical constraints of GPU memory. Preliminary tests indicated that batch sizes exceeding 50 were unfeasible for our single-GPU (NVIDIA RTX 3090) training setup. Consequently, we established the experimental batch size range for performance evaluation as [4, 8, 16, 32].

We evaluated performance metrics (MRR, Hit@1, and Hit@10) across this range. Optimal performance was achieved with a batch size of 16, while a batch size of 8 also yielded strong and comparable results. A decline in performance was observed for the smallest batch size tested (4) and for the largest (32), relative to these optima. The reduced performance at a batch size of 4 is consistent with the known challenge of less stable gradient estimations, which can hinder effective convergence. These findings indicate that batch sizes of 8 and 16 offer the most favorable performance characteristics for this task.

Notably, smaller batch sizes inherently increase training duration. This increase is attributable not only to the higher frequency of parameter updates but also to the potential underutilization of parallel processing capabilities in hardware accelerators. Therefore, while our results provide guidance on optimal performance, we recommend that users adjust the batch size according to their available computational resources and desired training speed. This flexible approach ensures that model training remains practical and can be tailored to individual operational contexts, balancing performance with computational feasibility.

\subsection*{Impact of Embedding Dimension \(D\) }
We experimented with embedding dimensions ranging from 16 to 96. Performance shows a trend of initial improvement followed by a plateau or gradual decline, with the best results observed at dimension \(D\) = 32. This phenomenon indicates that a moderate embedding dimension is sufficient to capture the necessary semantic details without incurring over-parameterization or redundancy. 

Our findings emphasize the critical role of carefully tuning the embedding dimensions. Not only do they balance model expressiveness and robustness, but they also play a fundamental part in capturing the inherent semantic structure of the biomedical knowledge graph. Given that the CP decomposition is specifically designed to unveil latent semantic patterns in multi-relational data, the chosen embedding dimension directly affects how effectively these semantic details are preserved in the learned representations. Therefore, determining an optimal embedding size is essential to fully realizing the potential of our model, ensuring that the resulting embeddings are compact enough to faithfully reflect the global semantic structure hidden in the data.

\subsection*{Impact of Fusion Weight \(\lambda\) }
The fusion weight \(\lambda\) is a crucial hyperparameter in BioGraphFusion, determining the balance between structural propagation and global semantic embeddings within the scoring function. A well-calibrated \(\lambda\) is essential for optimizing model performance across different datasets and biomedical knowledge graph structures. To systematically investigate its impact, we conducted experiments by varying \(\lambda\) over the set \(\{0.3, 0.4, 0.5, 0.6, 0.7, 0.8\}\), evaluating model effectiveness using MRR, Hit@1, and Hit@10.

Our results reveal a clear trend: performance, as measured by MRR, Hit@1, and Hit@10, consistently improves as \(\lambda\) increases from 0.3, peaking at \(\lambda = 0.7\) (MRR 0.429, Hit@1 0.377, Hit@10 0.529). Beyond this point, increasing \(\lambda\) to 0.8 leads to a slight performance decline. This observed peak at \(\lambda = 0.7\) suggests an optimal calibration where the model achieves a harmonious integration of its diverse information sources—primarily structural patterns and semantic knowledge. At this value, these components achieve a synergy, contributing in the most balanced and effective manner to the model's final predictions. The subsequent performance dip beyond \(\lambda = 0.7\) indicates that higher values may disrupt this calibrated balance, leading to a less complementary interplay of the model's informational components and, consequently, suboptimal generalization.

Consequently, the parameter \( \lambda \) in our scoring function, \( \tilde{f}(q_e, q_r, q_a) = \lambda f(q_e, q_r, q_a) + (1-\lambda) \phi(q_e, q_r, q_a) \), should not be misconstrued as a simple toggle weighting 'pure' Graph Structure Propagation (GSP) against 'pure' Knowledge Embedding (KE). Instead, \( \lambda \) fine-tunes the contributions of two distinct yet deeply intertwined components to the final score: \( f(q_e, q_r, q_a) \), which represents the output of our GSP module (itself initialized and guided by KE), and \( \phi(q_e, q_r, q_a) \), which is the direct score from our tensor decomposition-based KE module.

Given this inherent coupling---where KE directly informs the GSP process that yields \( f \)---KE's influence is integral to both terms, albeit differently. While \( \phi \) offers a direct semantic score, \( f \) provides structural insights shaped by KE. Thus, \( \lambda \) balances their ultimate contributions to the score, but the underlying structural information (channeled through \( f \)) and the overarching semantic context (from \( \phi \) and also embedded within \( f \)) are both fundamental to the model's predictive capabilities. Optimizing \( \lambda\) is therefore aimed at an ideal balance between global semantic knowledge from embeddings and the information gathered through semantically-guided graph propagation. This crucial balance ensures global semantics effectively steer graph propagation, helping to capture fine-grained, local interactions. A well-calibrated model can then comprehensively represent diverse relationships within biomedical graphs, ultimately yielding robust and accurate predictive scores.

\subsection*{Impact of Propagation Steps \(\ell\) }

We evaluated the number of propagation steps \(\ell\) over \(\{4, 5, 6, 7, 8\}\). Performance improves with increasing \(\ell\), peaking at \(\ell = 6\) (with MRR of 0.429, Hit@1 of 0.377, and Hit@10 of 0.529), and then declines at \(\ell = 8\). This suggests that an optimal propagation depth exists: too few steps limit the model's ability to capture extended neighborhood information, whereas too many steps lead to over-smoothing, causing node representations to become overly similar and less discriminative. The results indicate that a propagation depth of 6 best balances information aggregation and preservation of distinct node features.

The above analysis is specifically for the Disease-Gene dataset. Nevertheless, we speculate that a similar trend regarding the parameter \(\ell\) will be observed on other datasets, namely the existence of an optimal propagation depth that balances information propagation and over-smoothing. However, the specific numerical value of the optimal \(\ell\) may vary depending on the characteristics of the practical dataset.

\section*{(11) SM11: Protein-Protein Interaction Network Analysis of Known and Predicted Melanoma-Associated Genes}
\begin{figure*}[htbp]%调节图片位置，h：浮动；t：顶部；b:底部；p：当前位置
\centering
% {\color{black!20}\rule{213pt}{37pt}}
\includegraphics[width=.99\textwidth]{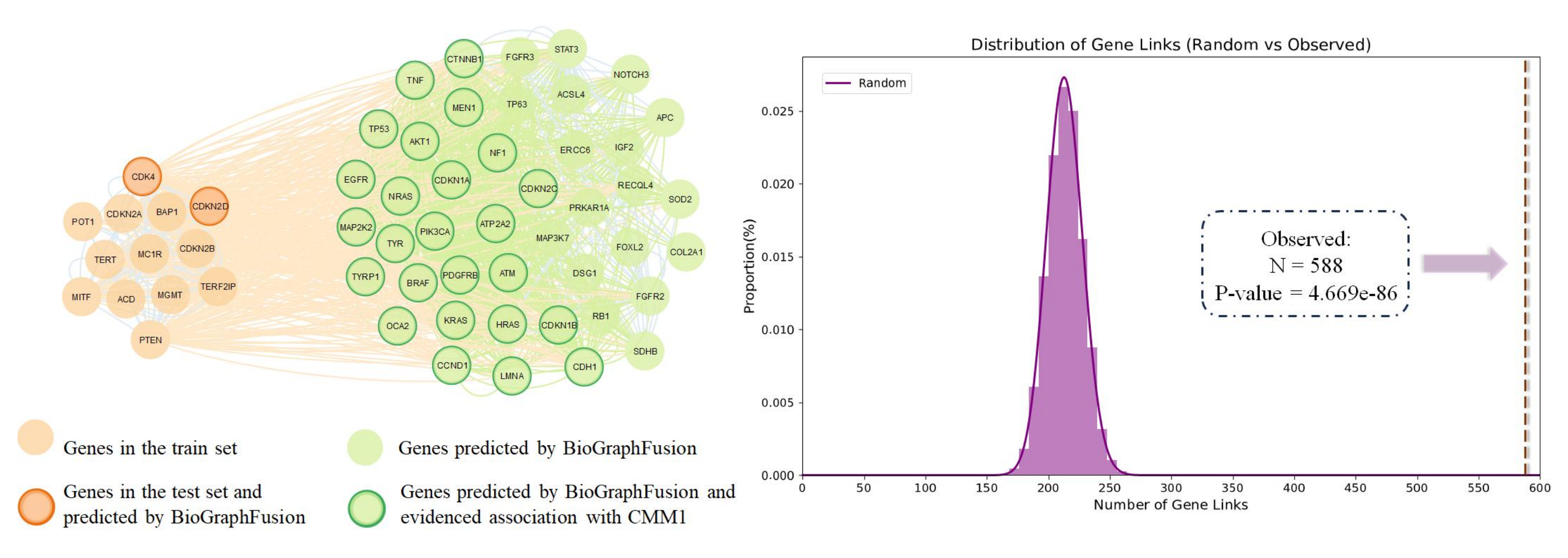}
\caption{Link visualization of known and predicted genes for melanoma on the PPI network. \& For melanoma, the observed number of network links is significantly larger than the random control (P = 4.669E-86, binomial test)}\label{fig:PPI&P-value} 
\end{figure*}

To assess the functional relevance of the genes identified by BioGraphFusion, we analyzed the connectivity patterns within the Protein-Protein Interaction (PPI) network. By comparing the observed number of interactions between known melanoma-associated genes and predicted candidates with a randomly expected baseline, we evaluated the statistical significance of these connections using a binomial test. The results clearly indicate that the observed connectivity is far greater than expected by chance.

The resulting PPI network (Fig.~\ref{fig:PPI&P-value}) reveals 588 actual interactions among known and predicted genes, substantially exceeding the expected 213.17 connections under a random model (P = 4.669E-86, binomial test). This high degree of interconnectivity suggests that the predicted genes are functionally linked to known melanoma-associated genes, potentially sharing key pathways involved in disease progression. These findings demonstrate the robustness and reliability of the BioGraphFusion predictive framework, reinforcing its potential as a powerful tool for uncovering novel candidate genes with functional relevance. The model’s ability to integrate diverse biological data and accurately predict gene interactions highlights its utility in advancing melanoma research and guiding further experimental validation.

% =========================
% 补充材料部分结束
% =========================

\end{document}